\documentclass[10pt,twocolumn,letterpaper]{article}
\usepackage{tikz}
\usetikzlibrary{spy}

\usepackage[normalem]{ulem}
\usepackage[pagenumbers]{cvpr} % To force page numbers, e.g. for an arXiv version

\usepackage{amssymb}
\usepackage{amsmath}
\usepackage{graphicx}
\usepackage{multirow} 
 \newcommand{\xiaoxu}[1]{#1}
 \newcommand{\zhongmin}[1]{#1}

\newcommand{\MethodNameShort}{DualPrim}

\newcommand{\netName}{DualPrim-Net}

\newcommand{\superquadric}{Q}
\newcommand{\superprimitive}{S}

\newcommand{\ax}{a_{x}}
\newcommand{\ay}{a_{y}}
\newcommand{\az}{a_{z}}

\newcommand{\epsa}{{\epsilon}_{1}}
\newcommand{\epsb}{{\epsilon}_{2}}

\newcommand{\rot}{R}
\newcommand{\trans}{T}

\newcommand{\cbasic}{c_{basic}}
\newcommand{\transparency}{\alpha}
\newcommand{\sharpness}{\theta}

\newcommand{\maskRender}{M_{render}}
\newcommand{\maskGT}{M_{gt}}

\newcommand{\imageRender}{I_{render}}
\newcommand{\imageGT}{I_{gt}}

\newcommand{\normalRender}{N_{render}}
\newcommand{\normalPred}{N_{Pred}}

\newcommand{\pe}{P_{E}}

\newcommand{\clr}{\textbf{c}}

\definecolor{OrangeColor}{rgb}{1.0,0.4,0}
\definecolor{DeltaColor}{rgb}{0.039,0.73,0.71}
\definecolor{SetaColor}{rgb}{0.867, 0.0235, 0.376}
\definecolor{SigmaColor}{rgb}{0.98,0.45,0.0}
\definecolor{RedColor}{rgb}{0.8,0,0}
\definecolor{AlphaColor}{rgb}{0,0,0.8}
\definecolor{BetaColor}{rgb}{0.8,0,0.8}
\definecolor{GammaColor}{rgb}{0.5,0,0.7}
\definecolor{EpsilonColor}{rgb}{0.353,0.725,0.906}
\definecolor{TauColor}{rgb}{0.423,0.235,0.192}

\newcommand{\cameraReady}[1]{\textcolor{OrangeColor}{#1}}
\renewcommand{\cameraReady}[1]{{#1}}

\newcommand{\immeshstype}{0.09}

\definecolor{cvprblue}{rgb}{0.21,0.49,0.74}
\usepackage[pagebackref,breaklinks,colorlinks,allcolors=cvprblue]{hyperref}

\def\paperID{***} % *** Enter the Paper ID here
\def\confName{CVPR}
\def\confYear{2026}

\title{DualPrim: Compact 3D Reconstruction with Positive and Negative Primitives}

\author{Xiaoxu Meng$^{1}$\thanks{Equal contribution}\\
{\tt\small xiaoxumeng.cs@gmail.com}
\and
Zhongmin Chen$^{4, 5}$\footnotemark[1]\\
{\tt\small chenzhongmin19@mails.ucas.ac.cn}
\and
Bo Yang$^{2}$\\
{\tt\small yangbo@waymo.com}
\and
Weikai Chen$^{1}$\thanks{Corresponding author, this paper solely reflects the author's personal research and is not associated with the author's affiliated institution.}\\
{\tt\small chenwk891@gmail.com}
\and
Weixiao Liu$^{3}$\\
{\tt\small  weixiaoliu@lucidmotors.com}
\and
Lin Gao$^{4, 5}$\\
{\tt\small gaolin@ict.ac.cn}
\and
$^{1}$Independent Researcher~
$^{2}$Waymo LLC~
$^{3}$Lucid Motors\\
$^{4}$Institute of Computing Technology, Chinese Academy of Sciences\\
$^{5}$School of Advanced Interdisciplinary Science, University of Chinese Academy of Sciences\\
}

\begin{document}
% \maketitle
\twocolumn[{%
    \renewcommand\twocolumn[1][]{#1}%
    \maketitle
    \vspace{-10mm}
    \begin{center}
    \centering
    \captionsetup{type=figure}
    \includegraphics[width=0.97\textwidth]{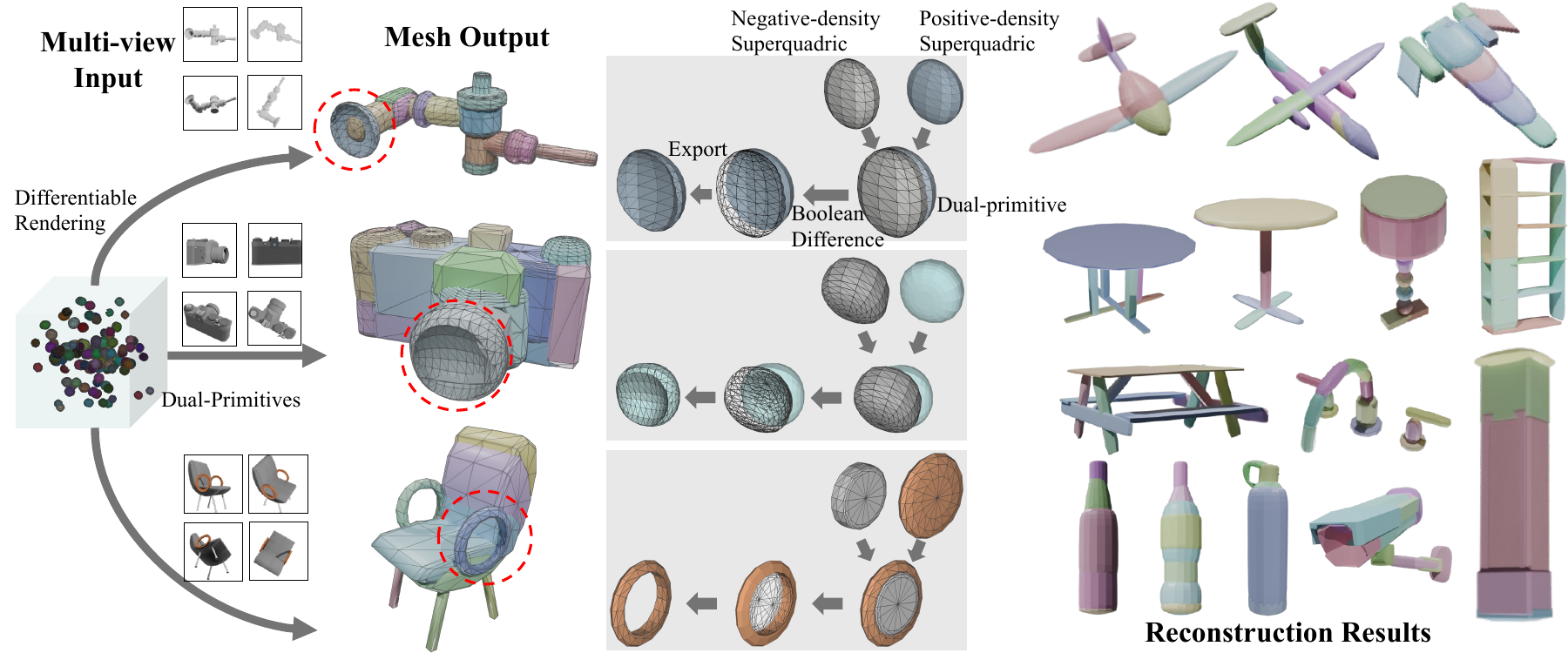}
    \captionof{figure}{
    \MethodNameShort~is a differentiable representation that models 3D shapes using \textit{dual-primitives} composed of positive- and negative-density superquadrics. This additive-subtractive design increases the representational power (e.g. holes and concavities) without sacrificing compactness or differentiability. Integrated into a volumetric differentiable renderer, \MethodNameShort~supports end-to-end learning from multi-view images and enables seamless mesh extraction through closed-form Boolean differencing. Our method produces compact, structured, and interpretable reconstructions that better meet downstream requirements than additive-only approaches.
    }
    \label{fig:teaser}
    \end{center}%
}]

\renewcommand{\thefootnote}{\fnsymbol{footnote}}
\footnotetext[1]{Equal contribution.}
\footnotetext[2]{Corresponding author, this paper solely reflects the author's personal research and is not associated with the author's affiliated institution.}
\vspace{-1mm}

% From: Single view or Multiview?
% From: Normal Image
\vspace{-8mm}
\begin{abstract}
Neural reconstructions often trade structure for fidelity, yielding dense and unstructured meshes with irregular topology and weak part boundaries that hinder editing, animation, and downstream asset reuse. We present DualPrim, a compact and structured 3D reconstruction framework. Unlike additive-only implicit or primitive methods, DualPrim represents shapes with positive and negative superquadrics: the former builds the bases while the latter carves local volumes through a differentiable operator, enabling topology-aware modeling of holes and concavities. This additive-subtractive design increases the representational power without sacrificing compactness or differentiability. We embed DualPrim in a volumetric differentiable renderer, enabling end-to-end learning from multi-view images and seamless mesh export via closed-form boolean difference. Empirically, DualPrim delivers state-of-the-art accuracy and produces compact, structured, and interpretable outputs that better satisfy downstream needs than additive-only alternatives.   
\end{abstract}

\vspace{-5mm}
\section{Introduction}
\label{sec:intro}

\begin{figure}[h!]
    \vspace{-2mm}
    \begin{minipage}[t]{\immeshstype\textwidth}
        \centering
        {\includegraphics[width=0.9\textwidth]{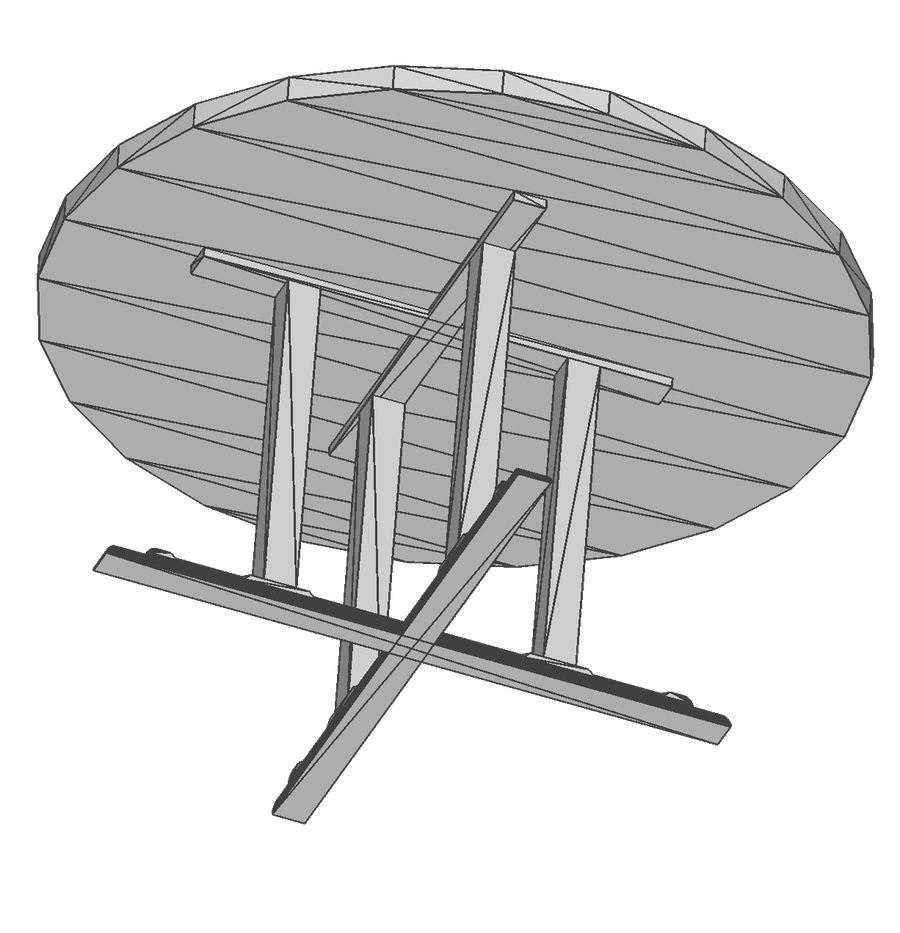}}
    \end{minipage}
    \begin{minipage}[t]{\immeshstype\textwidth}
        \centering
        % \subfloat[$\mathcal{L}_{max}=0$]
        {\includegraphics[width=1.0\textwidth]{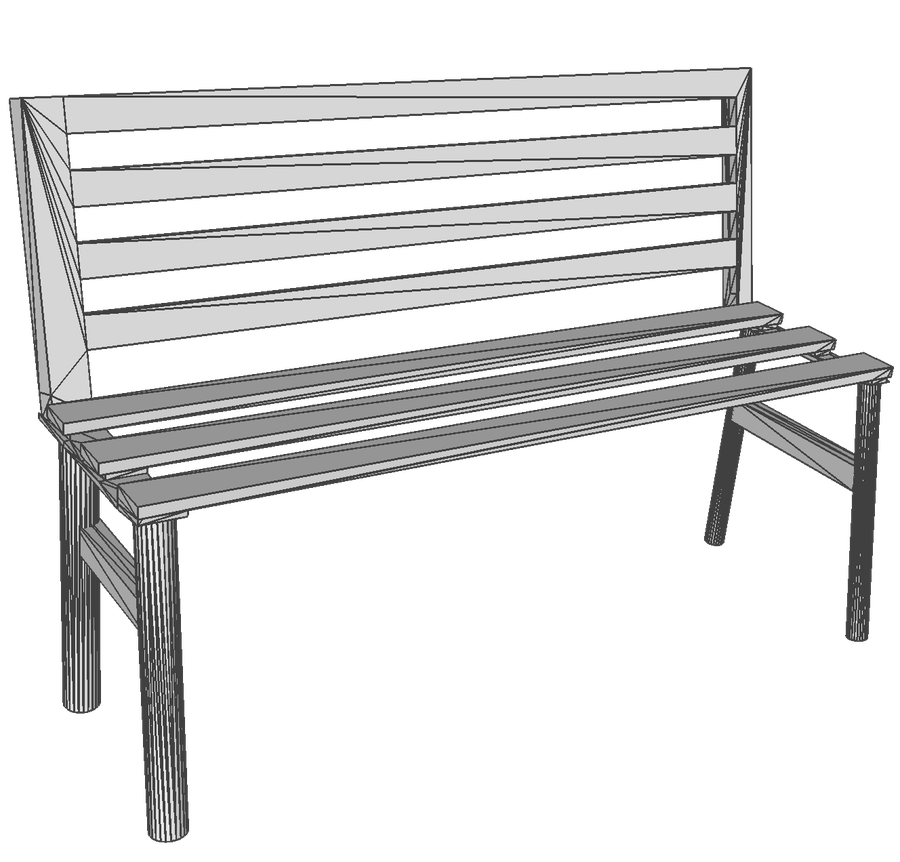}}
    \end{minipage}
    \begin{minipage}[t]{\immeshstype\textwidth}
        \centering
        % \subfloat[Baseline]
        {\includegraphics[width=0.9\textwidth]{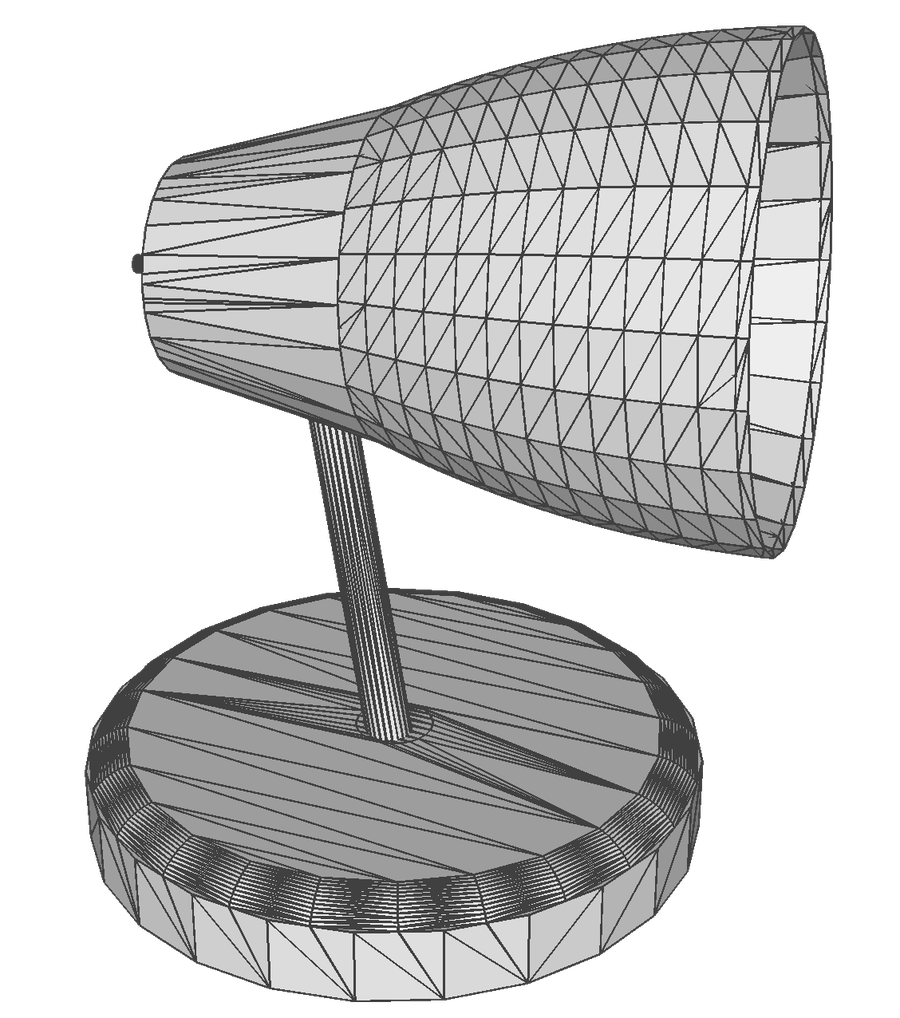}}
    \end{minipage}
    \begin{minipage}[t]{\immeshstype\textwidth}
        \centering
        % \subfloat[Baseline]
        {\includegraphics[width=1.0\textwidth]{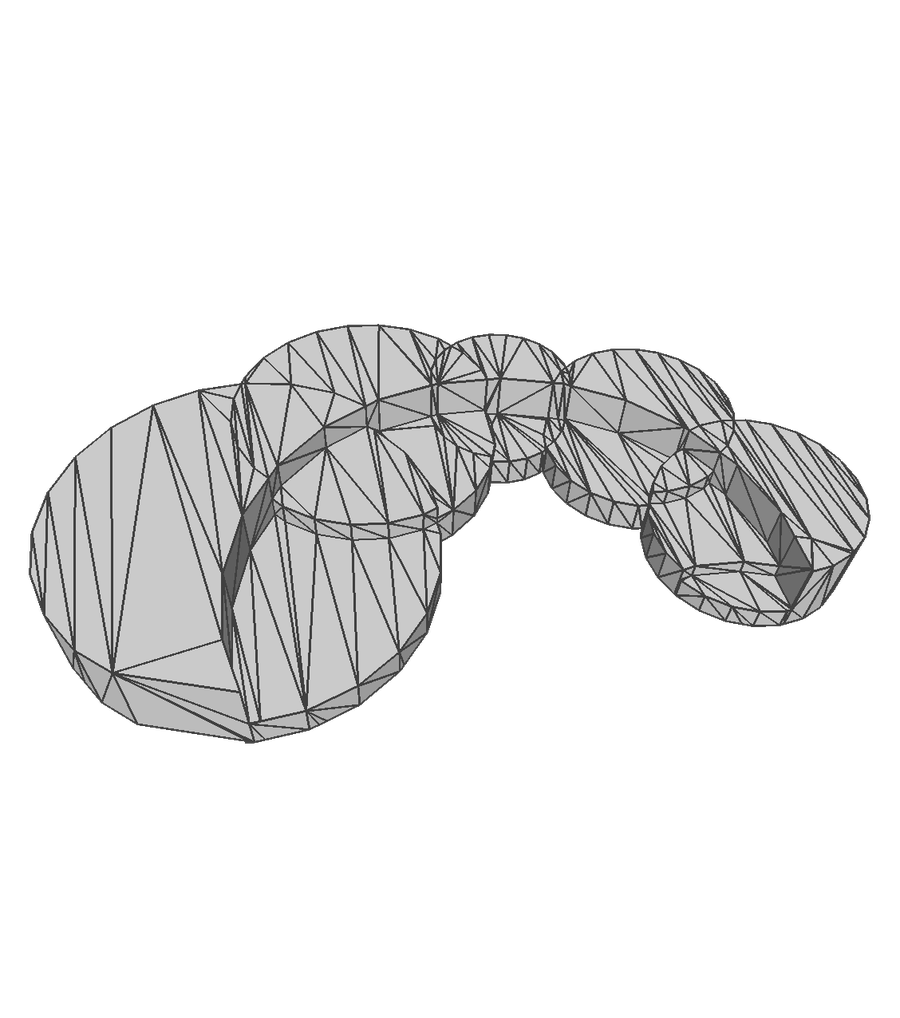}}
    \end{minipage}
    \begin{minipage}[t]{\immeshstype\textwidth}
        \centering
        % \subfloat[Baseline]
        {\includegraphics[width=0.7\textwidth]{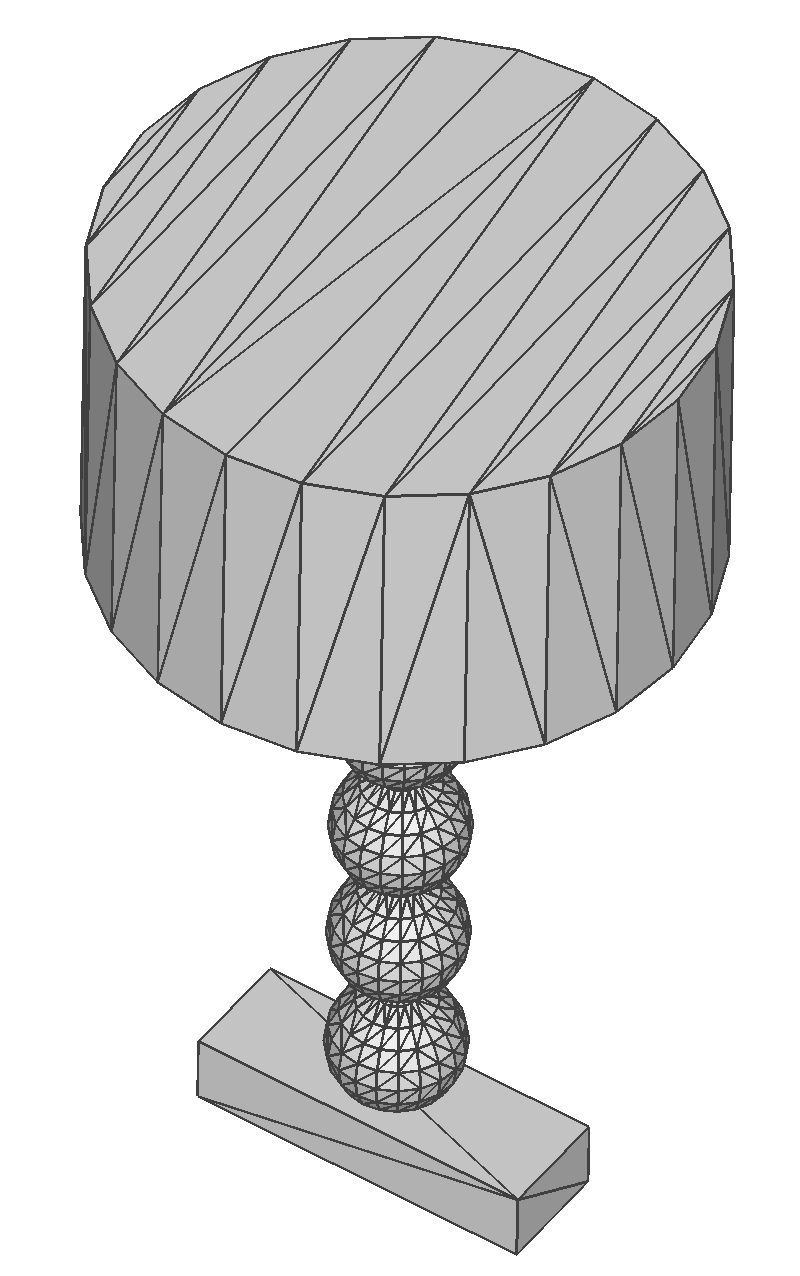}}
    \end{minipage}
    \\
    \begin{minipage}[t]{\immeshstype\textwidth}
        \centering
        {\includegraphics[width=0.9\textwidth]{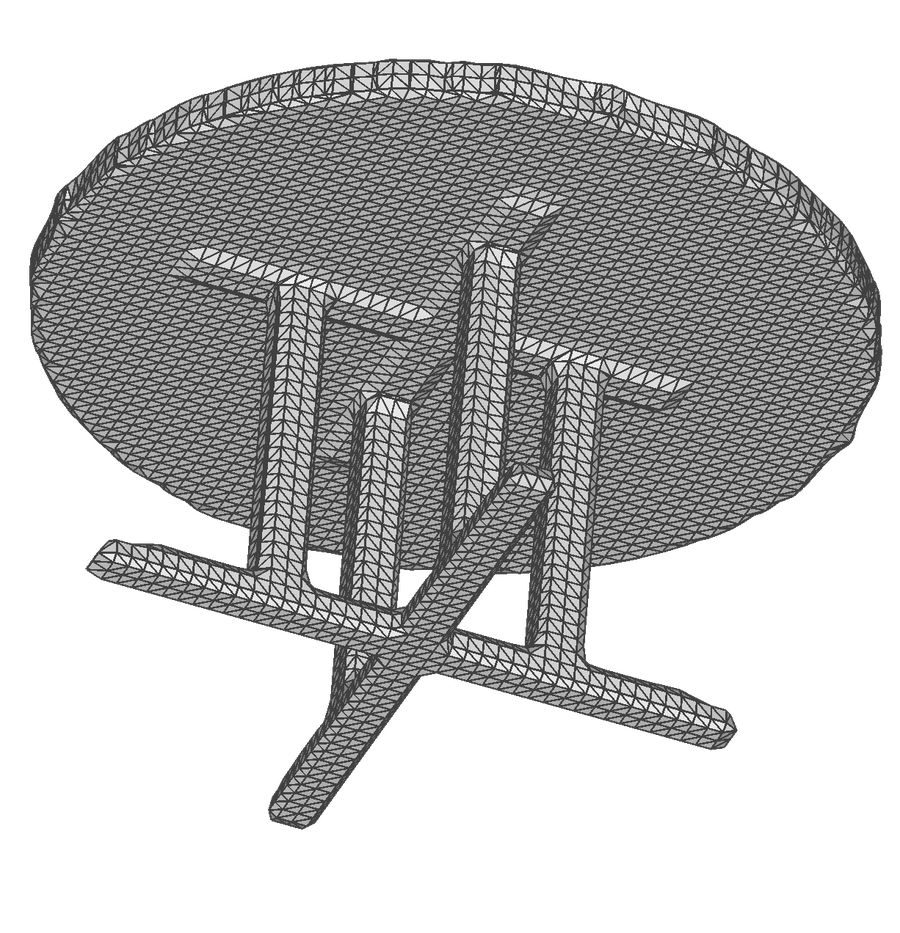}}
    \end{minipage}
    \begin{minipage}[t]{\immeshstype\textwidth}
        \centering
        % \subfloat[$\mathcal{L}_{max}=0$]
        {\includegraphics[width=1.0\textwidth]{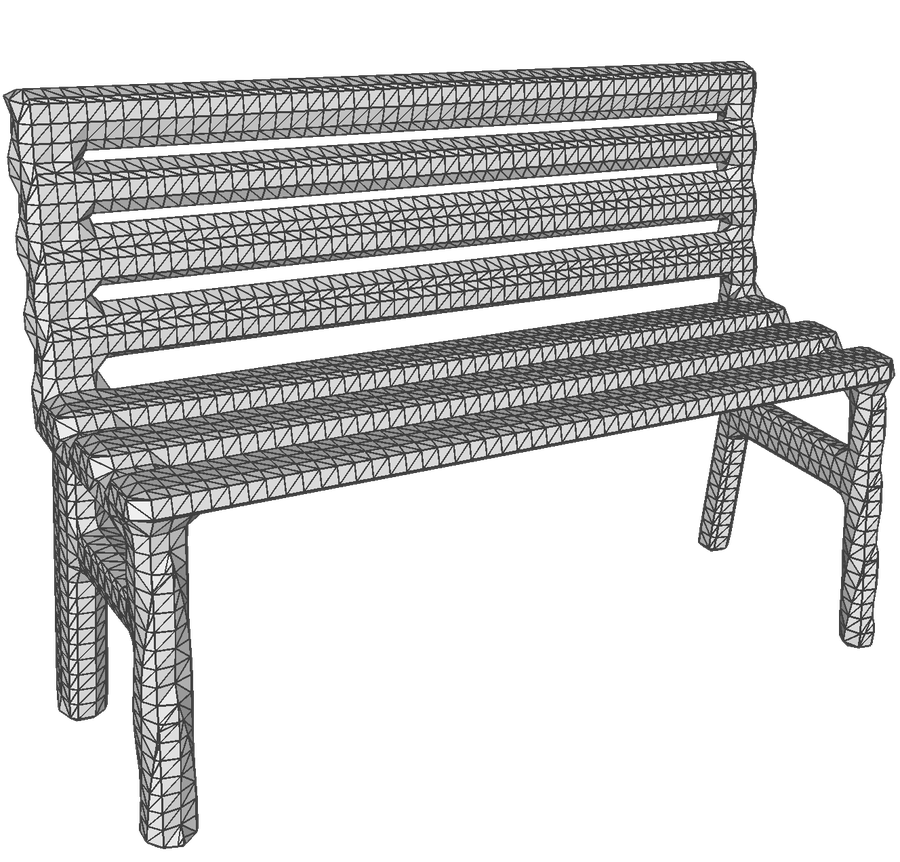}}
    \end{minipage}
    \begin{minipage}[t]{\immeshstype\textwidth}
        \centering
        % \subfloat[Baseline]
        {\includegraphics[width=0.9\textwidth]{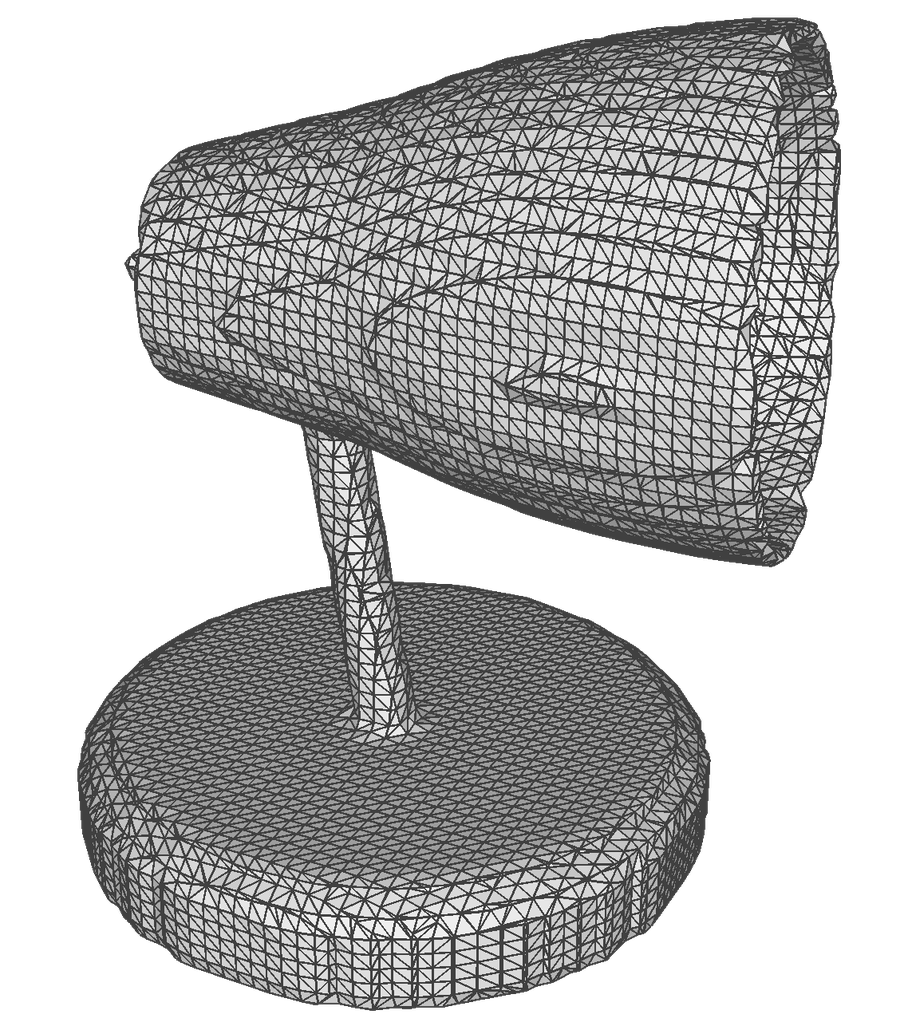}}
    \end{minipage}
    \begin{minipage}[t]{\immeshstype\textwidth}
        \centering
        % \subfloat[Baseline]
        {\includegraphics[width=1.0\textwidth]{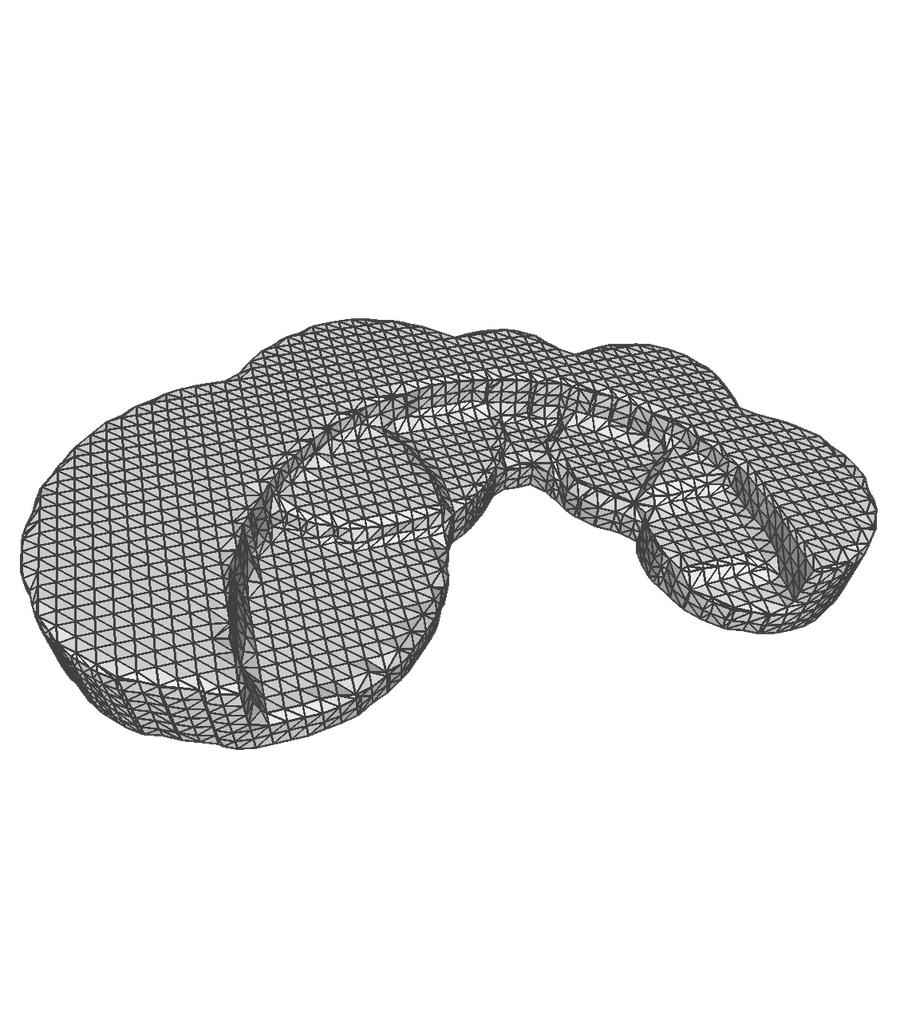}}
    \end{minipage}
    \begin{minipage}[t]{\immeshstype\textwidth}
        \centering
        % \subfloat[Baseline]
        {\includegraphics[width=0.7\textwidth]{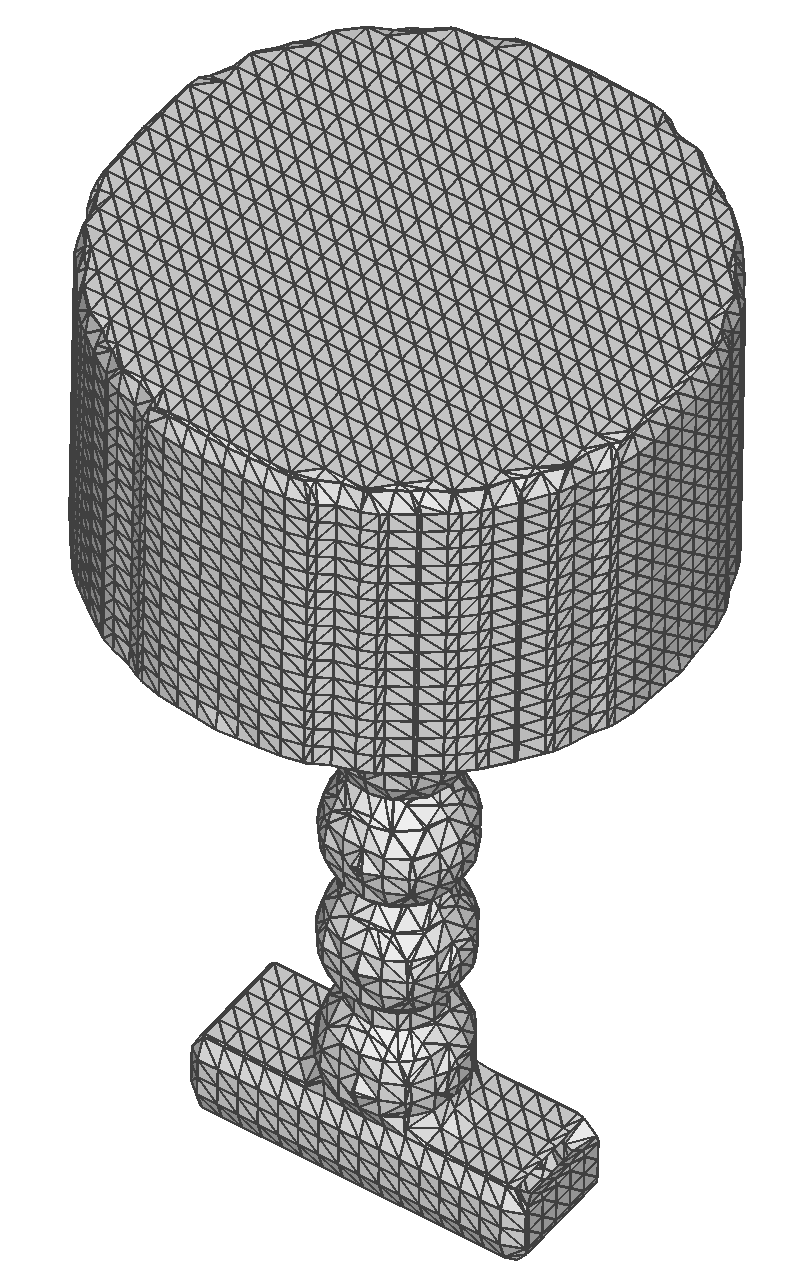}}
    \end{minipage}
    \vspace{-2mm}
    \caption{The first row shows the artist-created meshes (AMs), and the second row shows the exportation from Marching Cubes~\cite{marching_cubes}.
    }
    \vspace{-2mm}
    \label{fig:mesh_style_comparison}
\end{figure}

\vspace{-2mm}
Recent advances in neural reconstruction, such as NeRFs~\cite{mildenhall2020nerf,Sabour_2023_CVPR,Yang2023FreeNeRF}, SDFs~\cite{yariv2020idr,wang2022hfneus,wang2021neus,neus2,rnbneus}, and Gaussian Splatting (3DGS)~\cite{kerbl20233d,chen2024survey,fei20243d,Wu_2024_CVPR,huang20242dgs,Yu_2024_CVPR,Chen_2024_CVPR}, have achieved high-fidelity geometry modeling from multi-view images. 
However, these implicit or point-based representations produce dense and unstructured surfaces that are difficult to edit, animate, or integrate into standard graphics pipelines.
Artists and designers, on the other hand, rely on \textit{compact}, \textit{structured}, and \textit{part-based} meshes that explicitly capture both geometry and topology, as illustrated in Figure~\ref{fig:mesh_style_comparison}. 
The comparison highlights a key distinction: while artist-created meshes exhibit clean edge loops and consistent topology that support editing and animation\zhongmin{~\cite{sumner2004deformation,baran2007automatic}}, meshes extracted from implicit fields via Marching Cubes~\cite{marching_cubes} are typically over-tessellated and irregular, lacking meaningful part boundaries or structural organization. 
This contrast emphasizes a persistent gap between dense neural reconstructions and structured, interpretable geometry\zhongmin{~\cite{mescheder2019occupancy,niemeyer2020differentiable,oechsle2021unisurf}}, motivating the need for representations that are compact, topologically regularized, and 
% semantically
\cameraReady{structurally}
meaningful.

Primitive-based reconstruction offers a promising step toward structured geometry by representing shapes as combinations of analytic primitives such as cuboids or superquadrics\zhongmin{~\cite{tulsiani2017learning,paschalidou2019superquadrics}}.
Yet, most existing approaches rely solely on additive composition\zhongmin{~\cite{tulsiani2017learning,paschalidou2019superquadrics,paschalidou2021neural}}, accumulating positive volumetric fields to approximate target geometry. 
This additive nature inherently limits the ability to model topologically rich structures such as holes or concavities, which are ubiquitous in real-world objects\zhongmin{~\cite{Mo2020StructEdit}}.

Inspired by the constructive workflows used by 3D artists -- where modeling involves both adding and carving volumes -- We introduce \MethodNameShort, a dual-primitive representation that pairs each positive superquadric with a negative one.
The negative primitive acts as a differentiable carving operator, allowing the model to subtract local volumes and refine fine-scale structures while maintaining analytical regularity.
This constructive-subtractive mechanism greatly expands the expressiveness of primitive-based representations without sacrificing compactness or differentiability.
As shown in Figure~\ref{fig:teaser}, each dual-primitive is formed by combining a positive density superquadric that defines the constructive base shape with a negative density superquadric that carves out local regions through a differentiable boolean difference.
This allows \MethodNameShort~to model a wide variety of geometries, from smooth convex shapes to intricate topologies with internal cavities or open parts, using only a compact set of analytic parameters.

To train such a representation, we formulate \MethodNameShort~within a differentiable rendering framework\zhongmin{~\cite{niemeyer2020differentiable,oechsle2021unisurf}} that jointly optimizes the parameters of positive and negative superquadrics from multi-view image supervision. Along each camera ray, we integrate the volumetric contributions of all primitives, combining additive and subtractive density fields in a continuous, differentiable manner.
This unified formulation allows \MethodNameShort~to recover fine-scale geometry directly from 2D observations, while naturally supporting mesh extraction through boolean difference operations\zhongmin{~\cite{hertz2022spaghetti}}.
Together, these design choices enable \MethodNameShort~to bridge the gap between dense neural reconstructions and structured 3D geometry.
It produces meshes that are not only accurate and compact but also structurally organized, offering meaningful part decomposition and regular topology suitable for downstream applications such as editing, animation, simulation, and asset design\zhongmin{~\cite{Mo2020StructEdit}}.

\vspace{-2mm}

We evaluate on diverse multi-view reconstruction benchmarks covering a wide range of challenging 3D shapes.
Our method consistently outperforms state-of-the-art approaches both qualitatively and quantitatively.
Our main contributions can be summarized as:
\begin{itemize}
    \item A new dual-primitive representation that pairs positive superquadrics with negative ones, enabling differentiable volume carving. This constructive-subtractive formulation improves topological expressiveness (holes, concavities) while preserving compactness and analyticity.
    % that {reconstructs} 3D shapes with {structured topology,} regularized wireframes and easy conversion to artist-style triangular meshes. 
     % \yb{I feel it is not correct about ``easy conversion to triangular meth'', as marching cubes is easy but not usable}
    \item A novel differentiable rendering framework that jointly learns additive and subtractive primitives from multi-view images in an end-to-end manner and and supports direct mesh extraction via differentiable boolean operations.
    \item \MethodNameShort~establishes state-of-the-art reconstruction results with compact and structured geometry.
\end{itemize}

\section{Related Work}

\begin{figure*}[h!]
\vspace{-4mm}
\centering
    \includegraphics[width=0.9\linewidth]{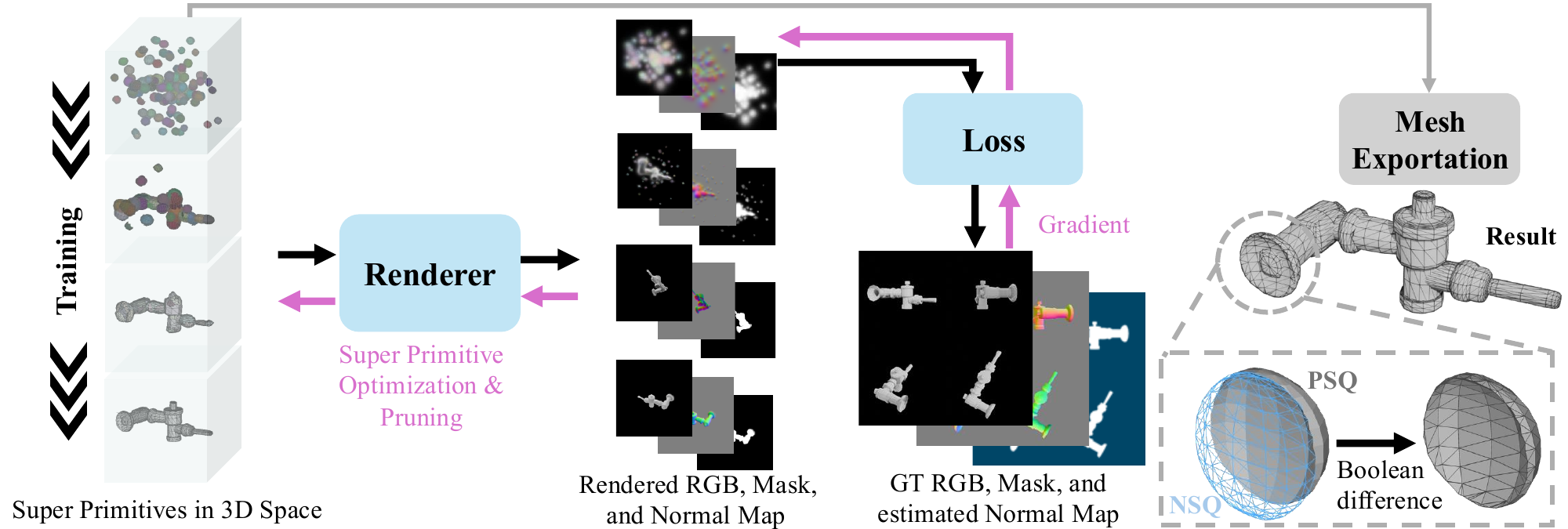}
\caption{The overview of \MethodNameShort~pipeline. We represent the 3D scene as a set of primitive parameters and optimize them within a differentiable rendering framework under multi-view image supervision. The final mesh is obtained by computing the Boolean difference between the positive and negative superquadric components of each primitive.
}
\vspace{-4mm}
% \yb{I feel the left three parts (sampled pixel, SDF cube, f(p,S_i) should have images as backgrounds; otherwise, no clues of what they are. Also, primitive pruning is not mentioned.}
\label{fig:pipeline}
\end{figure*}

\label{sec:related}
\paragraph{Neural Implicit Reconstructions.}
Recent advances in 3D reconstruction leverage neural implicit and explicit representations to recover geometry from multi-view images~\cite{sun2024recent}.
NeRF-variants~\cite{mildenhall2020nerf,Sabour_2023_CVPR,Yang2023FreeNeRF}
% ,muller2022instantngp,yang2024structurednerf,yariv2023multiview
optimize a continuous volumetric radiance field to produce high-fidelity novel views by fitting a per-scene neural field to input images. However, these methods do not provide explicit, compact surface representations, and rely on dense sampling and volumetric integration for geometry.
% SDF-based methods~\cite{yariv2020idr,wang2021neus,wang2022hfneus, rnbneus} 
\zhongmin{SDF-based and hybrid SDF methods~\cite{yariv2020idr,wang2021neus,wang2022hfneus, rnbneus, peng2025gaussian} including approaches that incorporate geometric priors for sparse-view reconstruction~\cite{mu2023neural},}
and Tetrahedrons-based methods~\cite{Munkberg_2022_CVPR,treece1999regularised}
improve surface fidelity by representing geometry as learned SDFs or tetrahedral grids and adapting volume rendering to reduce geometric bias. 
\zhongmin{Flexible isosurface extraction~\cite{shen2023flexicubes} further enables gradient-based mesh optimization by improving differentiability and geometric consistency during implicit-to-explicit conversion.}
While they achieve accurate zero-level surfaces, these methods inherit two key limitations: dependence on the implicit representation accuracy and sensitivity of subsequent mesh extraction to discretization and noise.
% Splatting~\cite{kerbl20233d,fei20243d,Chen_2024_CVPR,Wu_2024_CVPR}.
Overall, these approaches produce photorealistic, dense surfaces, but their underlying representations are typically unstructured and lack part-level interpretability, limiting their applicability for tasks that require semantic understanding or structural manipulation.
\zhongmin{In contrast, explicit surface reconstruction methods generate watertight and feature-preserving meshes from 3D point clouds, such as optimal-transport-based mesh reconstruction for CAD models~\cite{ye2024watertight}, but they are not differentiable and cannot be learned directly from image supervision.}
\vspace{-4mm}
\paragraph{Reconstruction from Model Retrieval and Constructive Methods.}
\zhongmin{Alternatively, structured 3D shapes can be reconstructed via retrieval or alignment to pre-existing CAD models~\cite{8954339,9009492,avetisyan2020scenecad,Gumeli_2022_CVPR,10.1145/3658236,caddeform,uy-joint-cvpr21}, though such approaches rely heavily on dataset priors.}
% Compact, structured, and interpretable 3D shapes can also be obtained through model retrieval and alignment approaches~\cite{8954339,9009492,avetisyan2020scenecad,Gumeli_2022_CVPR,10.1145/3658236,caddeform,uy-joint-cvpr21}. 
Constructive Solid Geometry (CSG) represents complex objects as hierarchical compositions of primitives via boolean operations. Neural CSG methods such as CSGNet~\cite{sharma2018csgnet}, UCSG-Net~\cite{kania2020ucsg}, \cameraReady{BSP-Net~\cite{chen2020bspnet}}, CSG-Stump~\cite{ren2021csg}, \cameraReady{CAPRI-Net~\cite{yu2022capri}} and D$^{2}$CSG~\cite{yu2023d} learn to parse or generate symbolic CSG trees.
However, neither model retrieval nor CSG-based approaches are differentiable, thereby hindering direct learning from images. Furthermore, hierarchical CSG parsing introduces combinatorial complexity: the search space of boolean trees grows exponentially with the number of primitives, which limits scalability to simple or low-part-count shapes.
Earlier works explored learning shape composition from elementary primitives for interpretable 3D generation and matching~\cite{deprelle2019learning}, 
as well as \cameraReady{convex decomposition to approximate shapes as unions of convex parts~\cite{deng2020cvxnet}, improving fidelity while maintaining part interpretability, but its convex parts are not explicitly aligned with semantic structure.}

% In contrast, our approach performs differentiable boolean reasoning directly between a set of positive and negative primitives, allowing end-to-end optimization from image supervision while preserving geometric interpretability.
\vspace{-4mm}
{\paragraph{Superquadric Reconstruction via Differentiable Rendering.}
Superquadrics have a long history in geometric modeling~\cite{barr1981superquadrics,pentland1987perceptual}. Recent works~\cite{liu2022EMS,wu2022primitive,Liu2023marching,kim2024tsqnet,fedele2025superdec} revisit them for robust, differentiable shape abstraction. Methods such as Iterative Superquadric Recomposition~\cite{alaniz2023iterative} and Differentiable Blocks World~\cite{monnier2023differentiable} optimize primitive parameters from 2D images, while GaussianBlock~\cite{jiang2024gaussianblock} integrates Gaussian fields for editable decomposition.
\cameraReady{Neural Parts~\cite{paschalidou2021neural} learns implicit part-based representations that enable semantic consistency and part-level manipulation. However, it does not enforce explicit geometric primitives.}
Beyond image-based learning, EMS~\cite{liu2022EMS} and Marching-Primitives~\cite{Liu2023marching} extract structured superquadric representations directly from 3D data, offering compact and interpretable models. Most prior works, however, rely on additive composition and cannot capture holes or subtracted regions. 
\cameraReady{
PrimitiveAnything~\cite{ye2025primitiveanything} extends this paradigm by formulating shape generation as an auto-regressive primitive assembly task.
}
\cameraReady{
DualVector~\cite{dualvector} leverages Boolean operations to produce editable vector representations, but it is restricted to 2D and lacks explicit primitive-based geometric regularization.
}
Our approach addresses this by incorporating both positive and negative superquadrics, enabling differentiable modeling of additive and subtractive relationships for expressive shape abstraction.
}
\vspace{-4mm}
\paragraph{Artist-Created Mesh Generation.}
\zhongmin{Complementary to reconstruction, recent generative works aim to mimic artist-created mesh design.}
% Beyond geometric reconstruction, recent generative models explore mesh synthesis mimicking human artistry.
MeshGPT~\cite{siddiqui2023meshgpt} and PolyGen~\cite{nash2020polygen} generate meshes autoregressively from a learned geometric vocabulary, while MeshAnything~\cite{chen2024meshanything,chen2024meshanythingv2artistcreatedmesh} converts 3D assets into artist-style meshes using transformer-based architectures.
Our approach differs in that it performs part-aware reconstruction directly from multi-view images rather than unconditional mesh generation, combining interpretability with geometric fidelity.

\section{Method}
\begin{table*}[t]
    \centering
    \begin{tabular}{|c|c|c|c|}
        \hline
        \textbf{Parameter Name} & \textbf{Shape} & \textbf{Explanation} & \textbf{Range} \\
        \hline
        scale parameters $(\ax, \ay, \az)$ & $(2, 3)$ & The radius of the x-, y-, z-axis of PSQ and NSQ. & $[0.02, 1.0]$\\
        \hline
        shape parameters $(\epsa, \epsb)$ & $(2, 2)$ & The parameters controlling the shape of PSQ and NSQ. & $[0.05, 2]$\\
        \hline
        primitive transparency $\transparency$ & $(1)$ & The transparency of the super-primitive. & $[0, 1]$\\
        \hline
        primitive render sharpness $\sharpness$ & $(1)$ & The sharpness of the super-primitive. & $[0, 1]$\\
        \hline
        translation $\trans$ & $(2, 3)$ & The translation of the PSQ and the NSQ. & $[-1, 1]$\\
        \hline
        rotation $\rot$ & $(2, 3)$ & The rotation angles of the PSQ and the NSQ. & $[-180, 180)$\\
        \hline
        basic color $\cbasic$ & $(3)$ & The basic RGB color of the super-primitive. & $[0, 1]$\\
        \hline
    \end{tabular}
    \caption{The full parameter list of a super primitive.}
    \vspace{-2mm}
    \label{table:primitive_property}
\end{table*}

% Xiaoxu: The following is copied from NEAT. need to make it different
% Given $N$ images ${\imageGT (k)}_{k=1}^N$ with a resolution of $(W,\ H)$ together with corresponding camera intrinsics, extrinsics, and object masks ${\maskGT (k)}_{k=1}^N$, 
% our goal is to reconstruct the 3D mesh of the object with our super-primitive representation. 

We propose a super-primitive based rendering framework that integrates both positive and negative shape components for compact 3D representation. %{leverage the volume rendering framework to render an object.}
The overall pipeline of our method is illustrated in Figure~\ref{fig:pipeline}. For each pixel on an input image, we sample $N$ points along the camera ray %emitting from the pixel
$\{\mathbf{p}(t_{i})=\mathbf{o} + t_{i}\mathbf{v}~|~i=1,2,...,N\}$,
where $\mathbf{o}$ denotes the camera center and $\mathbf{v}$ is the corresponding view direction. 
We implicitly represent the surface as a signed distance field (SDF) $f(\mathbf{p}(t_{i})): \mathbb{R}^{3}\rightarrow \mathbb{R}$. 
By accumulating the SDF-based densities $\sigma$ and colors $\clr_{i}$ of the sampled points, the predicted image $\imageRender$ is computed as:
%we can compute the predicted image $\imageRender$ with Equation~\ref{equ:pred_imgs}.

% \begin{equation}
%     \label{equ:pred_imgs}
%     \begin{aligned}
%         &\maskRender(\mathbf{o}, \mathbf{v}) = \int_{0}^{+\infty}\wt dt,
%         \\
%         &\imageRender(\mathbf{o}, \mathbf{v}) = \int_{0}^{+\infty}\wt \clr(\pt(t)))dt,
%         \\
%         &\normalRender(\mathbf{o}, \mathbf{v}) = \int_{0}^{+\infty}\wt f'(\pt(t)))dt.
%     \end{aligned}
% \end{equation}
\vspace{-4mm}
\begin{equation}
    \begin{aligned}
        &\imageRender(\mathbf{o}, \mathbf{v}) = \sum_{i=1}^{N}\big(\prod_{j=1}^{i-1}(1-\alpha_{j})\big)\alpha_{i} \clr_{i} ,
    \end{aligned}
    \label{equ:pred_imgs}
\end{equation}

\noindent where $\alpha_{i} = 1 - \exp(-\sigma_{i}\delta_{i})$. Here, $\delta_{i} = t_{i+1} - t_{i}$ is the distance between adjacent samples. 
To render a full image, 
% we simply compute the SDF values and auxiliary features derived from the super-primitive-based geometric representation.
we just need to calculate the SDF and the auxiliary features from the dual-primitive-based geometric representation.
\subsection{Formulation}

\label{sec:formulation}
% \zhongmin{Each super-primitive is represented by two superquadrics defined via the implicit equation:}
We represent each dual-primitive with two superquadrics defined by the implicit equation:
$$
f(x, y, z) = 
    ((
        (\frac{x}{\ax})^{\frac{2}{\epsb}}
        + 
        (\frac{y}{\ay})^{\frac{2}{\epsb}}
    )^\frac{\epsb}{\epsa}
    +
    (\frac{z}{\az})^{\frac{2}{\epsa}})^{\frac{\epsa}{2}} = 1.
$$

% \yb{Better to add all example shapes figure to visualize what are super-quadrtics.} Xiaoxu: Added in Figure 3
The first superquadric serves as a positive shape generator, referred to as the “positive density superquadric” (PSQ), while the second acts as a subtractive component, referred to as the “negative density superquadric” (NSQ).
%Here, the first super-quadrtic has positive density, referred to as the ``positive superquadric" (PSQ), while the second super-quadrtic has negative density, referred to as the ``negative superquadric" (NSQ). 
As shown in Figure~\ref{fig:teaser}, the NSQ functions like an eraser; it is effective only when overlapping with the PSQ, where it removes density within the intersection region.
%it is only effective when it overlaps with the PSQ, resulting in the erasure of the PSQ's density to zero within the region of overlap. 
If there is no overlap, the NSQ has zero contribution to the mesh, and the dual-primitive degenerates into one PSQ only.
%If there is no overlap between the PSQ and the NSQ, a super-primitive retrogrades to a super-quadratic.
This design significantly expands the representational capacity, enabling asymmetric and complex shapes beyond the limitations of conventional superquadrics.
% In this way, we greatly increase the range of representability from only symmetrical super-quadratics to various shapes.

Since the wireframe of each superquadric can be represented as edge loops, we can extract the 3D mesh of a dual-primitive by computing the Boolean difference between the meshes of its PSQ and NSQ components.
%mesh representing the PSQ and the mesh representing the NSQ. 
% \yb{I feel this paragraph needs more explanations, better with figure, to illustrate what is ``regularized way''}.
To learn dual-primitives directly from 2D observations, we parameterize each primitive as summarized in Table~\ref{table:primitive_property}, and define a differentiable rendering process based on these parameters.
% To learn the super-primitives from 2D images, we define a set of primitive parameters, as shown in Table~\ref{table:primitive_property}, and outline the rendering process using these parameters.
% \yb{It may be better to have this paragraph in 3.1}

\subsection{Renderer}
\label{sec:renderer}
Based on Equation~\ref{equ:pred_imgs}, the central challenge in rendering is estimating an appropriate density field $\sigma$. To achieve this, we first estimate the \cameraReady{implicit surface function (ISF)} for each point under every primitive, and then derive the corresponding probability density function from the ISF and primitive parameters of the $K$ dual-primitives.
% According to Equation~\ref{equ:pred_imgs}, the key issue in the rendering process is to find an appropriate density $\sigma$. We split this task into two steps: firstly, we estimate the SDF for each point for each primitive; then, we build a probability density function to estimate volume density $\sigma$ from the SDF and other primitive parameters of the $K$ super-primitives.

\paragraph{\cameraReady{Implicit Surface} Estimation.}
For each dual-primitive $\superprimitive$, the \cameraReady{implicit surface function} $f(\textbf{p}, \superquadric)$ for each superquadric $\superquadric$ (either PSQ or NSQ) for each 3D point $\textbf{p}$ is defined as:

\vspace{-2mm}
\begin{equation}
    \begin{aligned}
    f(\textbf{p}, \superquadric) = 
    (
        &(
            (
                \frac{\textbf{p}_{x}'}{a_{x,\superquadric}}
            )^{\frac{2}{\epsilon_{2,\superquadric}}}
            + 
            (
                \frac{\textbf{p}_{y}'}{a_{y,\superquadric}}
            )^{\frac{2}{\epsilon_{2,\superquadric}}}
        )^\frac{\epsilon_{2,\superquadric}}{\epsilon_{1,\superquadric}}
        +
        \\
        &(
            \frac{\textbf{p}_{z}'}{a_{z,\superquadric}}
        )^{\frac{2}{\epsilon_{1,\superquadric}}}
    )^{\frac{\epsilon_{1,\superquadric}}{2}} - 1,
    \end{aligned}
    \label{equ:sdf_superquadrtic}
\end{equation}
\vspace{-2mm}

where $\textbf{p}'$ denotes the 3D coordinate in the local frame of the superquadric, obtained by:
% where $\textbf{p}'$ is the 3D coordinate in the local superquadric coordinate, we can calculate $\textbf{p}'$ from Equation~\ref{equ:p_prime}.
\vspace{-1mm}
\begin{equation}
    \textbf{p}' = \rot^{-1}_{\superquadric}(\textbf{p} - \trans_{\superquadric}).
    \label{equ:p_prime}
\end{equation}
\vspace{-3mm}

\noindent After obtaining the ISF for PSQ and NSQ, we compute the ISF of the combined dual-primitive by subtracting the contribution of the NSQ from the PSQ.

To conduct the subtraction, we estimate $\pe$, the probability that the NSQ is effective, as:

\vspace{-2mm}
\begin{equation}
    \small
    \pe(\textbf{p}, S) = \Phi(-\frac{f(\textbf{p}, PSQ) }{\sharpness_{S}(\textbf{p})} - \mu) \cdot \Phi(-\frac{f(\textbf{p}, NSQ)}{\sharpness_{S}(\textbf{p})} - \mu),
    \label{equ:p_erase}
\end{equation}
where $\Phi(x) = (1 + \exp(-x))^{-1}$ is the Sigmoid function, and $\mu$ is a small offset ensuring that zero-crossings are preserved. $\pe$ attains high values only when $\textbf{p}$ lies within both PSQ and NSQ and the primitive is sufficiently sharp. 
The merged ISF $f(\textbf{p}, \superprimitive)$ and normal $\textbf{n}(\textbf{p}, \superprimitive)$ are computed as:
\vspace{0mm}
\begin{equation}
    \small f(\textbf{p}, \superprimitive) = f(\textbf{p}, PSQ)\cdot
        (1 - \pe(\textbf{p}, \superprimitive)) - f(\textbf{p}, NSQ)\cdot\pe(\textbf{p}, \superprimitive),
    \label{equ:sdf_primitive}
\end{equation}

\vspace{-4mm}

\begin{equation}
    \begin{aligned}
        \textbf{n}(\textbf{p}, \superprimitive) =~&
        \textbf{Normalize}(f'(\textbf{p}, PSQ)) \cdot (1 - \pe(\textbf{p}, \superprimitive))\\
        -~&
        \textbf{Normalize}(f'(\textbf{p}, NSQ)) \cdot \pe(\textbf{p}, \superprimitive).
    \end{aligned}
    \label{equ:normal_primitive}
\end{equation}
\vspace{-2mm}

\begin{figure}[t!]
\centering
    \includegraphics[width=0.9\linewidth]{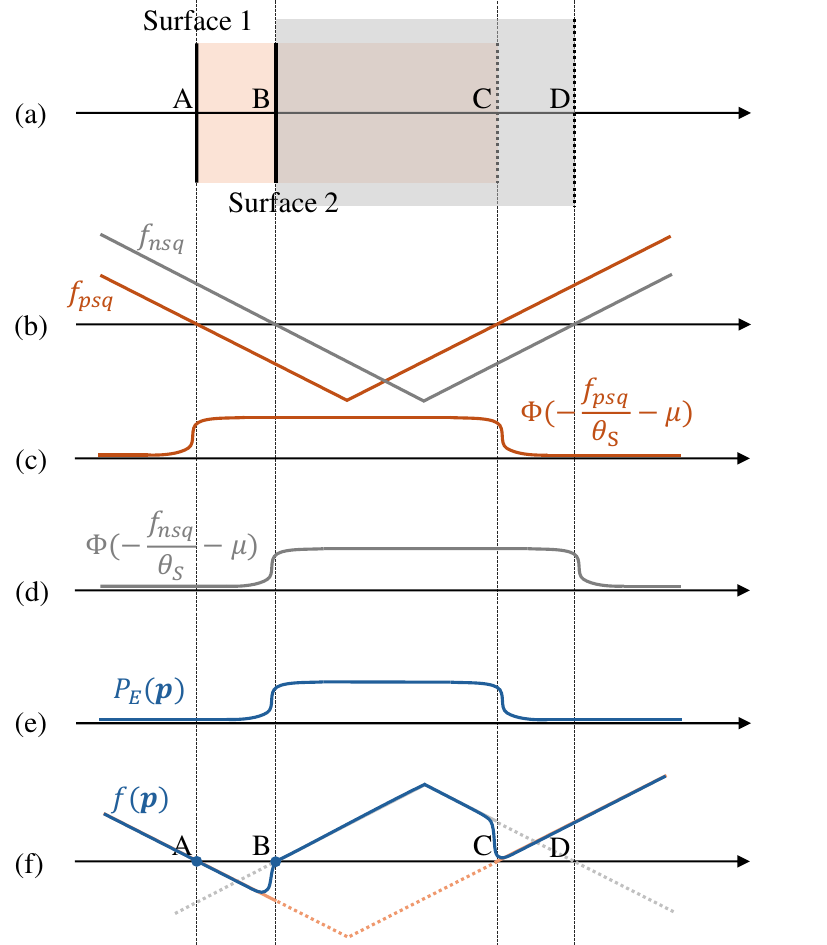}
\caption{Illustration of the SDF calculation with PSQ and NSQ. }
\vspace{-3mm}
% \yb{I would suggest adding numbering, e.g., a, b, c, ..., to each row, and refer them clearly in the corresponding text part. Current text is somewhat not easy to follow with the figure.}
\label{fig:illustratioin}
\vspace{-3mm}
\end{figure}

We provide a 2D illustration of this subtraction mechanism in Figure~\ref{fig:illustratioin}. The ray–surface intersections and the resulting ISF transitions demonstrate how NSQ effectively erases portions of PSQ, creating complex composite shapes through a simple probabilistic formulation.
% \reviseToConfirm{We show a 2D illustration of rendering a super-primitive with subtraction in Figure~\ref{fig:illustratioin}. As shown in Figure~\ref{fig:illustratioin} (a), the ray intersects with the PSQ on Point A and Point C, and intersects with the NSQ on Point B and Point D. As shown in Figure~\ref{fig:illustratioin} (e), $\pe = 1$ for the overlapping region between Point B and Point C. Point A is an intersection point for PSQ with $\pe(A) = 0$, $f(A) = f_{psq}(A) = 0$. The rendering of Point A is not affected by NSQ. For Point B, we have $\pe(B) = 1$ and $f(B) = -f_{nsq}(B) = 0$. The subtraction operation creates a new intersection for this super-primitive at point B. Point C is also an intersection point for PSQ. However, we have $\pe(C) = 1$ and $f(C) = -f_{nsq}(C) > 0$. Therefore, Point C is erased by NSQ and will not be rendered. For Point D, we have $\pe(D) = 0$ and $f(D) = f_{psq}(D) > 0$. This point will not be rendered.}

\vspace{-3mm}
\paragraph{Probability Density Estimation.}
Following NeuS~\cite{wang2021neus}, we estimate the volume density of each primitive from the ISF values as:
% with Equation~\ref{equ:density_primitive}.
\vspace{-3mm}

\begin{equation}
    \sigma_\superprimitive(\textbf{p}) = 
    \max \big(
        \frac{\Phi(
            \frac{f(
                \textbf{p} + \Delta \textbf{p}, \superprimitive
            )}{\sharpness_{\superprimitive}}
        ) - \Phi(
            \frac{f(
                \textbf{p} - \Delta \textbf{p}, \superprimitive
            )}{\sharpness_{\superprimitive}}
        )}{\Phi(
            \frac{
                f(\textbf{p} + \Delta \textbf{p}, \superprimitive
            )}{\sharpness_{\superprimitive}}
        )},
        0
    \big).
    \label{equ:density_primitive}
\end{equation}
\vspace{-2mm}

The final point-wise density is then obtained by summing over all $K$ dual-primitives:
% Then, to render the RGB image, we take the sum of all the $K$ super-primitives densities as the density of the point 
$\sigma(\textbf{p}) =
    \sum_{k=1}^{K}{\transparency_{\superprimitive_{k}}(\textbf{p})
    \cdot
    \sigma_{\superprimitive_{k}}(\textbf{p})}
$.
% We further model the point color $\textbf{c}(\textbf{p})$ as a combination of the primitive-specific appearance and a learned lighting field, as:
We consider that the color of the point is affected by the the texture and the lighting condition. We use Equation~\ref{equ:color_point} to calculate the color of the point. Here, $C(\textbf{p}) \in \mathbb{R}^{3}$ is the texture and lighting predicted with an MLP.
\vspace{-2mm}
\begin{equation}
    \textbf{c}(\textbf{p}) = 
        \sum_{k}
        {
            c_{basic}(\textbf{p}, \superprimitive_{k}) 
            \cdot 
            \frac{
                \sigma_{\superprimitive_{k}}(\textbf{p})
            }{
                \sigma(\textbf{p})
            }
        }
        + C(\textbf{p}),
    \label{equ:color_point}
\end{equation}
\vspace{-2mm}

Similarly, we render the mask $\maskRender(\mathbf{o}, \mathbf{v})$ and normal map $\normalRender(\mathbf{o}, \mathbf{v})$ using volume accumulation:
% We can render the mask $\maskRender(\mathbf{o}, \mathbf{v})$ along with the rendered image:
\vspace{-1mm}
\begin{equation}
    \label{equ:pred_mask}
    \begin{aligned}
        &\maskRender(\mathbf{o}, \mathbf{v}) = \sum_{i=1}^{N}\big(\prod_{j=1}^{i-1}(1-\alpha_{j})\big)\alpha_{i}.
    \end{aligned}
\end{equation}
\vspace{-3mm}

% Additionally, we can calculate the rendered normal with Equation~\ref{equ:pred_normal}. Here, the normal of each point $\textbf{n}_{i}$ can be calculated with Equation~\ref{equ:normal_point}.
\begin{equation}
    \label{equ:pred_normal}
    \normalRender(\mathbf{o}, \mathbf{v}) = 
    \sum_{i=1}^{N}
    \big(
        \prod_{j=1}^{i-1}(1-\alpha_{j})
    \big)
    \alpha_{i}\textbf{n}_{i},
\end{equation}
\vspace{-2mm}
where each $\textbf{n}_{i}$ is computed as:
\vspace{-1mm}
\begin{equation}
    \label{equ:normal_point}
    \textbf{n}(\textbf{p}) = 
    \sum_{k}
    {
        \textbf{n}_{\textbf{p},\superprimitive_{k}}(
            \textbf{p}, \superprimitive_{k}
        )
        \cdot
        \frac{
            \sigma_{S_{k}}(\textbf{p})
        }{
            \sigma(\textbf{p})
        }
    }.
\end{equation}
\vspace{-3mm}

\begin{figure}[h!]
\vspace{-3mm}
\centering
    \includegraphics[width=1.0\linewidth]{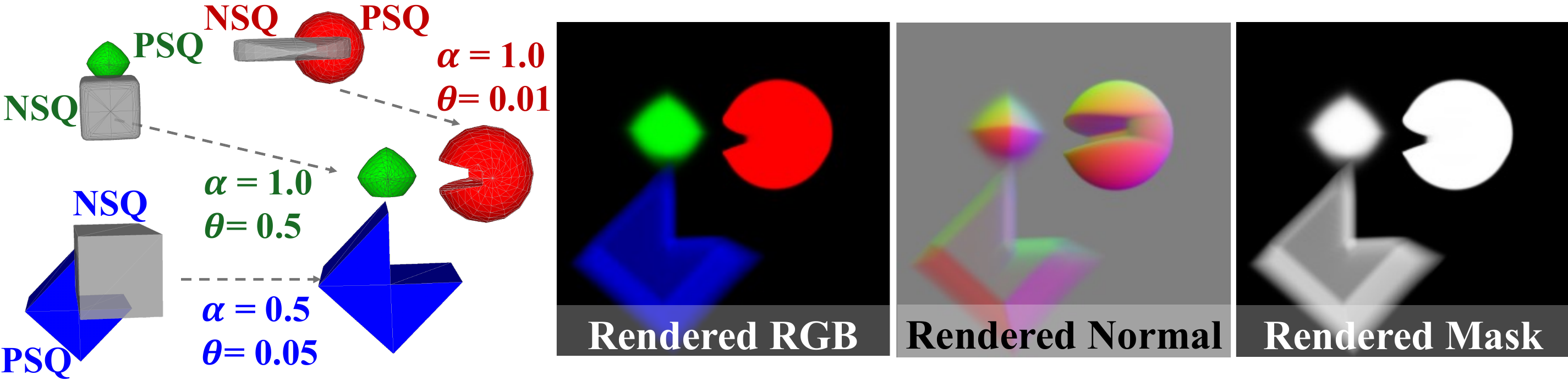}
\caption{We show the rendering of three example dual-primitives.
\vspace{-6mm}
}
% \yb{I feel the left three parts (sampled pixel, SDF cube, f(p,S_i) should have images as backgrounds; otherwise, no clues of what they are. Also, primitive pruning is not mentioned.}
\label{fig:vis_basic_dualprim}
\end{figure}

\noindent As demonstrated in Figure~\ref{fig:vis_basic_dualprim}, our renderer effectively synthesizes diverse primitives with controllable shape, color, transparency, and surface sharpness, enabling accurate 3D reconstruction and fine-grained structure representation.

% As shown in Figure~\ref{fig:vis_example}, our renderer is able to render primitives with different shape, color, transparency, and sharpness.
\vspace{-3mm}
\subsection{Mesh Exportation}
In the mesh extraction stage, we discard primitives with low transparency ($\transparency < T_{export}=0.5$) and export the meshes corresponding to both PSQs and NSQs.
The final object mesh is obtained by performing a Boolean difference operation between the PSQ and NSQ meshes, allowing precise geometric reconstruction of complex structures and maintaining structured and regularized wireframes.
% And we extract the 3D mesh of the object by computing the boolean difference between the meshes representing the PSQ and the meshes representing the NSQ.
\vspace{-1mm}
\section{Optimization}
\subsection{Training Objectives}
\label{sec:method_training}

We supervise the training of \netName{} using six loss terms, formulated as:
\vspace{-1mm}
\begin{equation}
    \begin{aligned}
        \mathcal{L}=&\mathcal{L}_{rgb}
               +\lambda_{mask}\cdot\mathcal{L}_{mask}
               +\lambda_{sparse}\cdot\mathcal{L}_{sp} 
               +\lambda_{e}\cdot\mathcal{L}_{e} \\
               &+\lambda_{max}\cdot\mathcal{L}_{max}
               +\lambda_{norm\_reg}\cdot\mathcal{L}_{norm\_reg}.
    \end{aligned}
\end{equation}
\vspace{-3mm}

The overall objective combines photometric, geometric, and structural regularizations to ensure faithful reconstruction and stable optimization.  
The first two terms correspond to photometric and mask supervision, respectively:
% The photometric {RGB Loss} and {mask Loss} are defined as
\begin{equation}
    \mathcal{L}_{rgb} = \sum_{i, j}||\imageRender(i,j) - \imageGT(i,j)||\cdot \maskGT(i,j),
\end{equation}
\vspace{-3mm}
\begin{equation}
    \mathcal{L}_{mask} = \sum_{i, j}BCE(\maskRender(i,j), \maskGT(i,j)).
\end{equation}
\vspace{-2mm}

\noindent 
Beyond direct supervision, we use 4 regularization terms to enhance structural compactness and geometric consistency.
% We use 4 regularization terms to enhance the reconstruction.

\vspace{-3mm}
\paragraph{Redundant Suppression.}
To discourage using redundant dual-primitives and promote compact surface, we impose a sparsity loss that penalizes excessive opacity:
% For real-world objects with open structures, the surfaces are sparsely distributed in the 3D space. To prevent the renderer from using redundant super-primitives, we introduce a sparsity regularization to promote the formation of surfaces:
\vspace{-2mm}
\begin{equation}
    \mathcal{L}_{sp} = \frac{1}{K}\sum_{\textbf{p}}\alpha(\textbf{p}).
\end{equation}
\vspace{-5mm}

\paragraph{Rendering Probability Entropy Regularization.}
Since an object's physical presence is inherently binary (either exists or not), the transparency $\alpha(\textbf{p})$ should ideally converge to 0 or 1.  
To encourage such deterministic behavior, we minimize the entropy of $\alpha(\textbf{p})$:
% In the physical world, the existence of object is binary (exist/not exist). As a result, the transparency of a primitive is either 0 (exist) or 1 (not exist). We therefore add the entropy of $\alpha(\textbf{p})$ as an extra regularization:

\vspace{-3mm}
\begin{equation}
    \small
    \mathcal{L}_{e} = -
    \frac{1}{K}\sum_{\textbf{p}} \alpha(\textbf{p})\log(\alpha(\textbf{p}))
    +
    (1 - \alpha(\textbf{p}))
    \log((1 - \alpha(\textbf{p}))).
\end{equation}

\vspace{-3mm}

\paragraph{Rendering Probability Maximum Value Regularization.}
Physically, the surface existence probability should not exceed 1.  
To enforce this constraint, we introduce a max-value regularization that softly penalizes $\alpha(\textbf{p}) > 1$:

% In the physical world, the probability of surface existence cannot exceed 1. As a result, we add the maximum alpha loss of $\alpha(\textbf{p})$ as an extra regularization:
\vspace{-2mm}
\begin{equation}
    \mathcal{L}_{max} = \frac{1}{K}\sum_{\textbf{p}} {ReLU(\alpha(\textbf{p}) - 1)}.
\end{equation}

\vspace{-5mm}
\paragraph{Normal Consistency Regularization.} 
To further encourage geometrically coherent reconstruction, we add a normal consistency term.  
This loss enforces agreement between the predicted surface normal $\normalRender(i,j)$ and a reference normal $\normalPred(i,j)$, which is estimated directly from the rendered surface without requiring additional annotations:
% To guide the network towards geometrically consistent prediction, we introduce a normal regularization loss, which is based on the predicted normal itself (without additional annotations). This loss encourages the model to produce more structured and coherent representations.
% We additionally use \textbf{normal loss} as extra supervision. The normal map $\normalPred(i,j)$ used as supervision is predicted with the state-of-the-art normal prediction approach StableNormal~\cite{ye2024stablenormal}.
\vspace{-2mm}
\begin{equation}
    \mathcal{L}_{norm\_reg} = \sum_{i, j}||\normalRender(i,j) - \normalPred(i,j)||\cdot \maskGT(i,j).
\end{equation}
\vspace{-2mm}

Together, these loss terms jointly guide the model to produce compact, physically plausible, and geometrically consistent reconstructions.
% \paragraph{Sharpness Regularization}
% In the physical world, the sharpness of the primitives is close to one. As a result, we add the sharpness regularization loss of $\theta(\textbf{p})$ as an extra regularization:
% \begin{equation}
%     \mathcal{L}_{sharpness} = \frac{1}{N_{sample}}\sum_{\textbf{p}} \theta(\textbf{p}).
% \end{equation}

\vspace{-1mm}
\subsection{Adaptive Control of Dual-primitives}
We initialize the reconstruction with a dense set of randomly distributed dual-primitives and progressively refine this set through adaptive pruning.  
This adaptive process iteratively removes redundant primitives and retains only those essential to accurately represent the target structure.

\vspace{-4mm}
\paragraph{Pruning Strategy.}
Following a strategy similar to 3DGS~\cite{kerbl20233d}, we use the primitive transparency parameter $\transparency$ as a key pruning criterion.  
Specifically, dual-primitives with $\transparency < 0.02$ are discarded, as they contribute negligibly to the final rendering.  
To avoid numerical instability and preserve structural clarity, we also remove primitives whose scale parameters fall below a threshold $t_{a} = 0.01$.  
This ensures that all retained primitives correspond to meaningful, spatially significant parts of the object.

\vspace{-4mm}
\paragraph{View-dependent Filtering.}
We prune primitives with negligible rendering weights across all viewpoints — typically those fully occluded or visually redundant.  
This pruning is performed periodically during training, allowing the model to dynamically reallocate representational capacity toward visible and structurally informative regions.

Through this adaptive control mechanism, the model evolves from a dense initialization to a compact, interpretable, and efficient primitive representation, leading to faster convergence and improved reconstruction quality.

% We begin with a random set of dense points and apply our adaptive method to iteratively prune the super-primitives. This process refines the initial dense set into a sparse configuration that accurately represents the object's structure. Similar to the approach used in Gaussian Splatting~\cite{kerbl20233d}, we use the primitive transparency $\transparency$ parameter as a key stopping criterion and prune the super-primitives with $\transparency < 0.02$. Additionally, super-primitives with scale parameters below a given threshold, $t_{a} = 0.01$, are pruned, ensuring that each remaining primitive corresponds to a sufficiently large portion of the object.
% We also prune the primitives that have very low rendering weights in any of the viewpoints, e.g., occluded in all views. The pruning is conducted periodically during the reconstruction process.

\renewcommand{\arraystretch}{1.1}
\newcommand{\shapenetImgWidth}{0.10}
\setlength{\tabcolsep}{10pt}

% --------------------------------------------------
% Macro for one image
% --------------------------------------------------
% \newcommand{\itwimg}[1]{\includegraphics[width=\itwImgWidth\textwidth]{#1}}
\newcommand{\shapenetImg}[2][]{%
  \vbox{%
    \hbox{\includegraphics[width=\shapenetImgWidth\textwidth]{figures/shapenet/#2}}%
    \ifx &#1&%
      % do nothing if #1 is empty
    \else
      % \vskip 1mm
      \vspace{-1mm}
      \hbox to \shapenetImgWidth\textwidth{\hss\small #1\hss}%
    \fi
  }%
}
% --------------------------------------------------
% Macro for one object (two rows)
% args:
% 1 object folder name
% 2 object name prefix
% 3 view id
% --------------------------------------------------
\newcommand{\twoRowShapenet}[2]{

% ---- first row ----
\shapenetImg[Input]{#1_input.png} &
\shapenetImg[Ours]{#2_#1_ours.png} &
\shapenetImg[PA~\cite{ye2025primitiveanything}]{#2_#1_pa.png} &
\shapenetImg[MA~\cite{chen2024meshanything}]{#2_#1_ma.png} &
\shapenetImg[MP~\cite{Liu2023marching}]{#2_#1_mp.png} &
\shapenetImg[EMS~\cite{liu2022EMS}]{#2_#1_ems.png} &
\shapenetImg[NVD~\cite{Munkberg_2022_CVPR}]{#2_#1_nvdiffrec.png}
\\

% ---- second row ----
\shapenetImg[FC~\cite{shen2023flexicubes}]{#2_#1_flexicube.png} &
\shapenetImg[RNS~\cite{rnbneus}]{#2_#1_neus2.png} &
\shapenetImg[{CAPRI}~\cite{yu2022capri}]{#2_#1_capri.png} &
\shapenetImg[{D$^{2}$CSG}~\cite{yu2023d}]{#2_#1_d2csg.png} &
\shapenetImg[{DiffCSG}~\cite{yuan2024diffcsg}]{#2_#1_diffcsg.png} &
\shapenetImg[2DGS~\cite{huang20242dgs}+MC]{#2_#1_2dgs_mc.png} &
\shapenetImg[2DGS~\cite{huang20242dgs}+CSG~\cite{yu2022capri}]{#2_#1_2dgs_csg.png}
\vspace{-1mm} \\[2pt]
}

\begin{figure*}[t]
\centering

\begin{tabular}{c c c c c c c}

% --------------------------------------------------
% Objects
% --------------------------------------------------

\twoRowShapenet{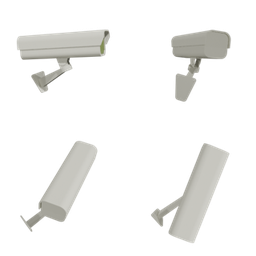}{camera}
\vspace{-3mm}
\twoRowShapenet{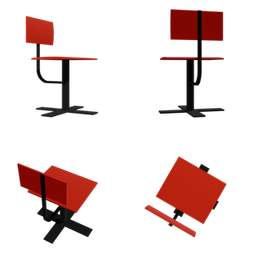}{chair}
\vspace{-3mm}
\twoRowShapenet{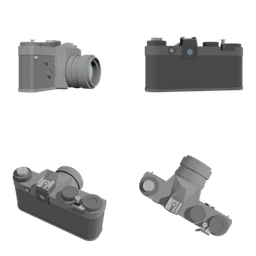}{camera}
\vspace{-3mm}
\twoRowShapenet{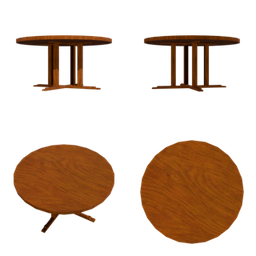}{table}
\vspace{-3mm}
\twoRowShapenet{2b82a928c4a81fe1df4cfe396cee719e}{chair}
\end{tabular}
\vspace{-2mm}
\caption{
Qualitative comparison on in-the-wild objects.
Each object occupies two rows.
The first row shows input, Ours, PA, MA, MP, EMS, and NVD.
The second row shows FC, RNS, CAPRI, D$^{2}$CSG, DiffCSG, 2DGS+MC, and 2DGS+CSG.
}
\vspace{-2mm}
\label{fig:comparison_shapenet}
\vspace{-2mm}
\end{figure*}
\section{Experiments}
\begin{figure*}[h!]
    \begin{minipage}[t]{0.12\textwidth}
        \centering
        \subfloat[Input]
        {\includegraphics[width=\textwidth]{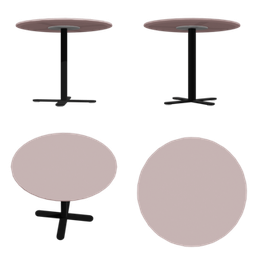}}
    \end{minipage}
    \begin{minipage}[t]{0.11\textwidth}
        \centering
        \subfloat[$\mathcal{L}_{sp}=0$]
        {\includegraphics[width=0.85\textwidth]{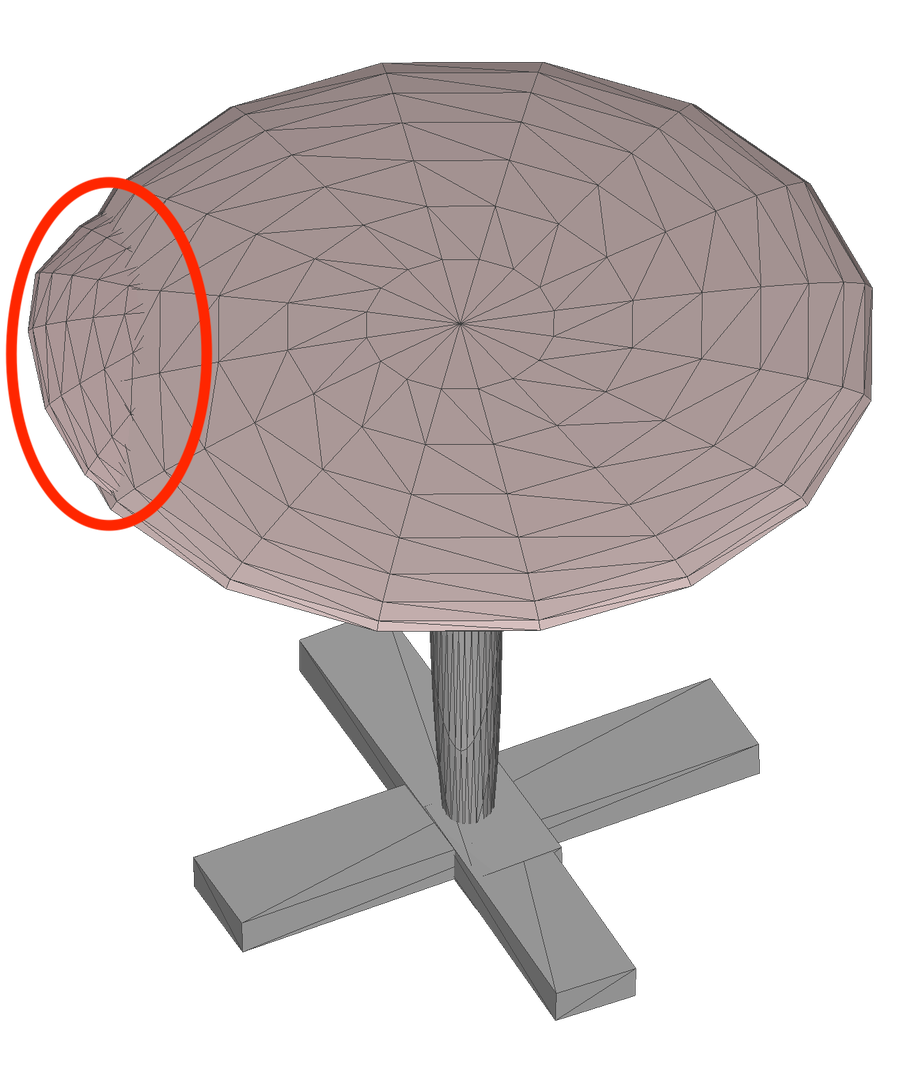}}
    \end{minipage}
    \begin{minipage}[t]{0.11\textwidth}
        \centering
        \subfloat[Ours]
        {\includegraphics[width=0.85\textwidth]{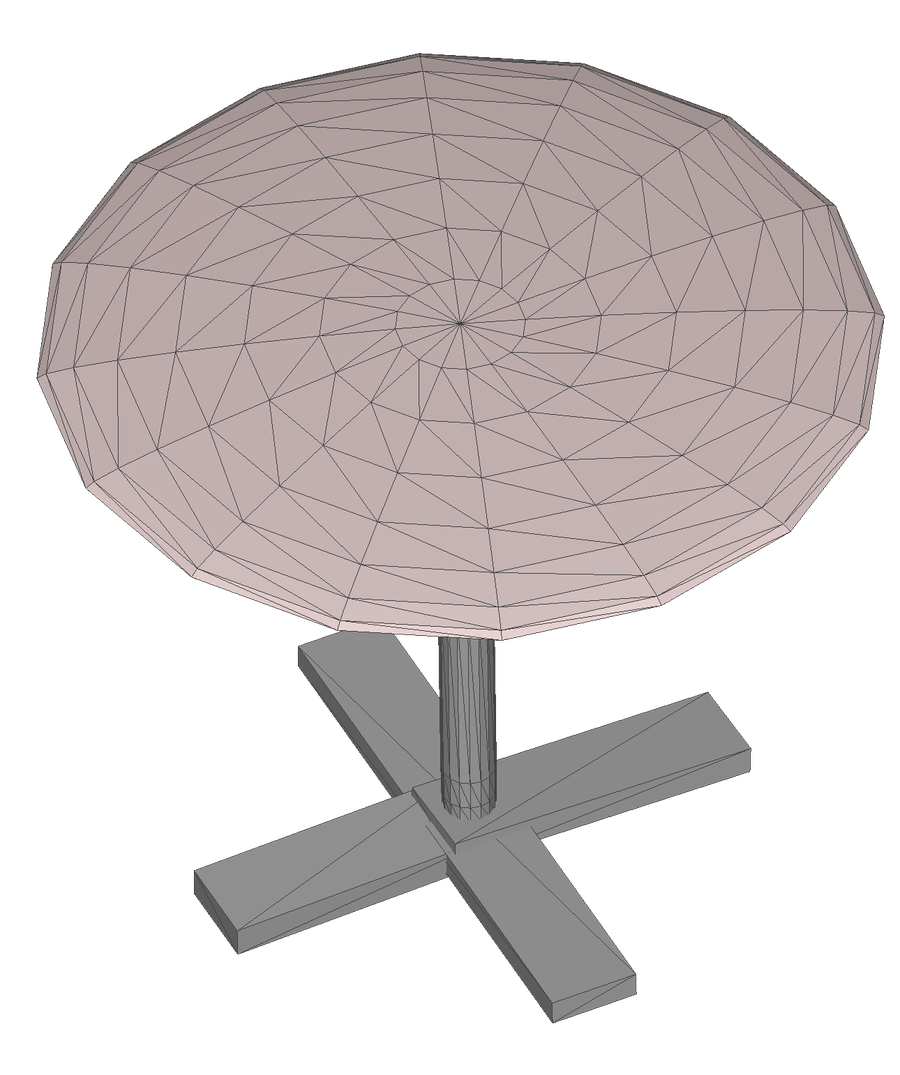}}
    \end{minipage}
    \begin{minipage}[t]{0.12\textwidth}
        \centering
        \subfloat[Input]
        {\includegraphics[width=\textwidth]{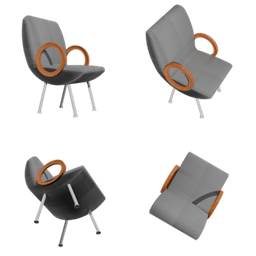}}
    \end{minipage}
    \begin{minipage}[t]{0.12\textwidth}
        \centering
        \subfloat[$\mathcal{L}_{max}=0$]
        {\includegraphics[width=0.8\textwidth]{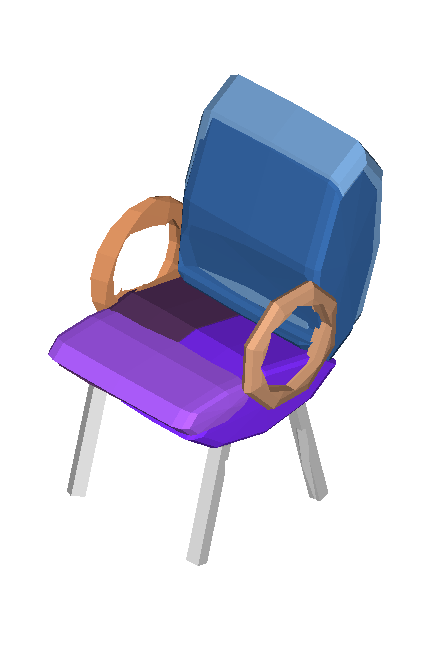}}
    \end{minipage}
    \begin{minipage}[t]{0.12\textwidth}
        \centering
        \subfloat[Ours]
        {\includegraphics[width=0.8\textwidth]{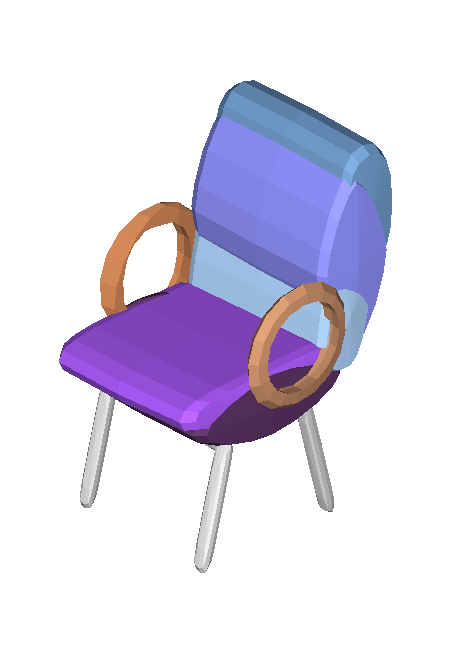}}
    \end{minipage}
    \begin{minipage}[t]{0.13\textwidth}
        \centering
        \subfloat[PSQ only]
        {\includegraphics[width=0.75\textwidth]{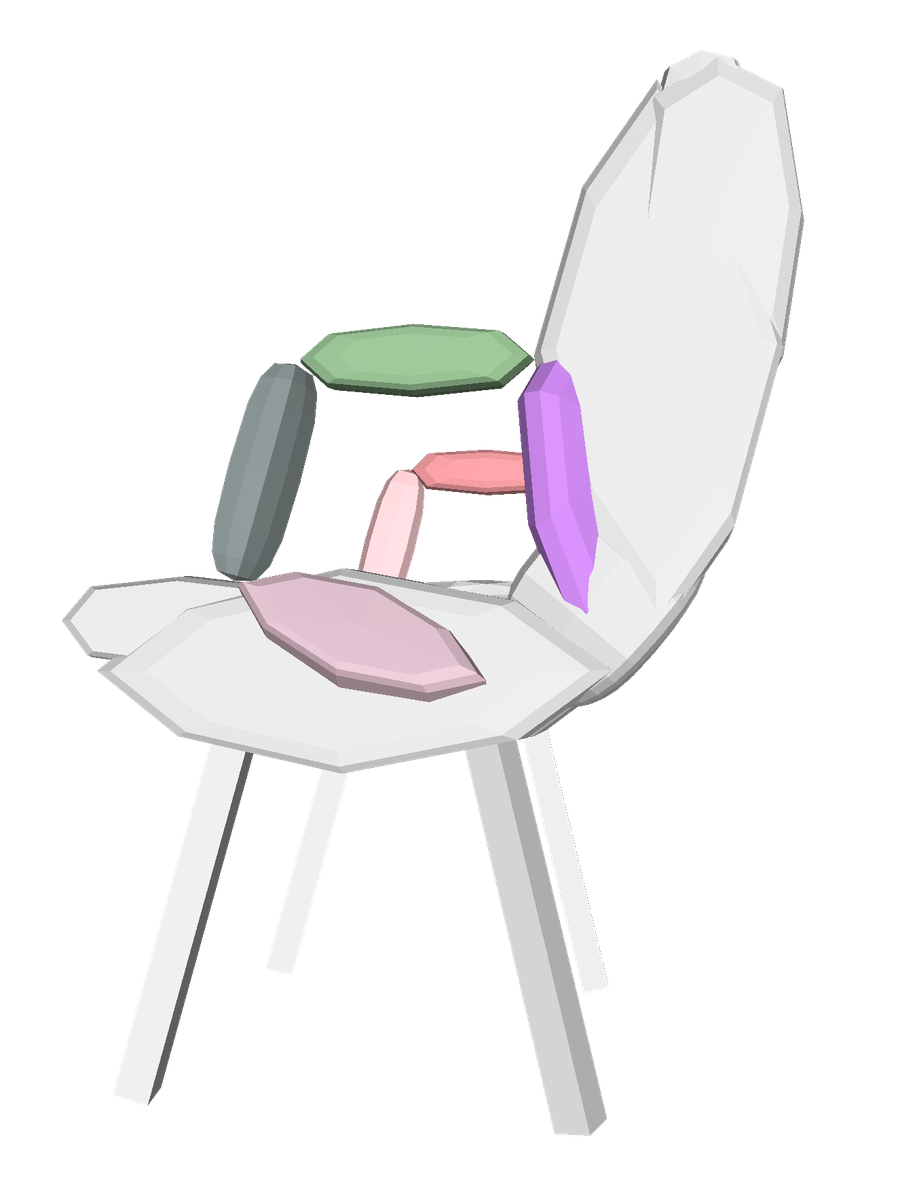}}
    \end{minipage}
    \begin{minipage}[t]{0.13\textwidth}
        \centering
        \subfloat[PSQ+NSQ]
        {\includegraphics[width=0.75\textwidth]{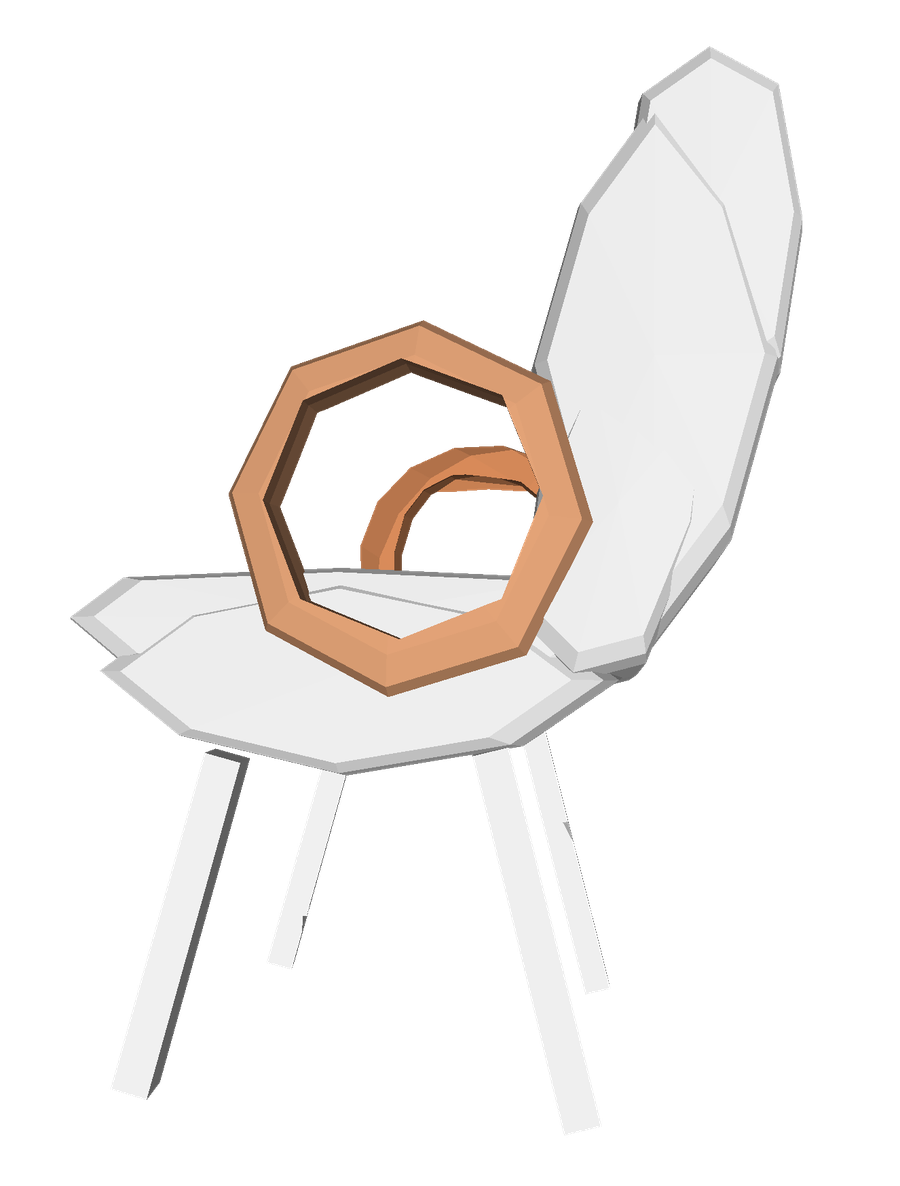}}
    \end{minipage}
    \vspace{-2mm}
    \caption{The ablation study on regularization terms.
    }
    \label{fig:ablation_regularization}
\vspace{-4mm}
\end{figure*}

% \input{figures/fig_ablation}

% \vspace{-2mm}
\subsection{Experimental Setup}
% \vspace{-2mm}
\paragraph{Datasets.}
\label{sec:experiment_setup}
\zhongmin{We evaluate \MethodNameShort~on multi-view reconstruction tasks to demonstrate its ability to achieve state-of-the-art reconstruction quality while maintaining a compact and regularized wireframe representation. }
% We evaluate \MethodNameShort~through multi-view reconstruction experiments to demonstrate that it achieves state-of-the-art reconstruction quality while maintaining a compact and regularized wireframe representation. 
\zhongmin{Experiments are conducted on $12$ categories from the ShapeNet dataset~\cite{shapenet2015}, with $15$ objects randomly sampled per category to cover a diverse range of textures and geometries. }
% The evaluation is conducted on $12$ categories from the ShapeNet dataset~\cite{shapenet2015}, with $15$ objects randomly sampled from each category, covering a diverse range of textures and geometries. 
\zhongmin{We uniformly sample $24$ viewpoints on the unit sphere, adding top and bottom views to obtain $26$ camera positions. }
% To generate the multi-view images, we uniformly sample $24$ viewpoints on the unit sphere and add the top-view and bottom-view as the camera positions, resulting in $26$ viewpoints.
\zhongmin{Ground truth 3D models are rendered using $Blender$ at a resolution of $256 \times 256$. }
% We render the ground truth 3D model with $Blender$ at a resolution of $256 \times 256$.
\zhongmin{All experiments are evaluated against representative state-of-the-art methods.}
% All experiments are compared with the SOTA methods for better verification.
\vspace{-2mm}
\paragraph{Implementation details.}
\zhongmin{We initialize $K=100$ dual-primitives randomly in the $[-1, 1]$ space. 
The MLP for lighting consists of $4$ layers with Xavier initialization.}
\vspace{-2mm}
\paragraph{Baselines.}
\zhongmin{We compare \MethodNameShort~with state-of-the-art methods across 5 categories: 
volume rendering-based method RNb-NeuS (RNS)~\cite{rnbneus},
\cameraReady{
Gaussian-based method 2DGS~\cite{huang20242dgs}
},
tetrahedron-based method nvdiffrec (NVD)~\cite{Munkberg_2022_CVPR}, 
% primitive-based differentiable rendering method Differentiable Blocks
% World (DBW)~\cite{monnier2023differentiable}, 
shape abstraction methods EMS~\cite{liu2022EMS}, Marching Primitives (MP)~\cite{Liu2023marching}, PrimitiveAnything (PA)~\cite{ye2025primitiveanything}, 
\cameraReady{
CapriNet (CAPRI)~\cite{yu2022capri}, D$^{2}$CSG~\cite{yu2023d},
DiffCSG~\cite{yuan2024diffcsg},
}
and topology optimization methods Flexicubes (FC)~\cite{shen2023flexicubes} and MeshAnything (MA)~\cite{chen2024meshanything}. }
% We compare our approach with the SOTA volume rendering-based method RNb-NeuS (RNS)~\cite{rnbneus}, tetrahedron-based method nvdiffrec (NVD) ~\cite{Munkberg_2022_CVPR}. We also compare with the SOTA primitive-based shape abstraction approach Marching Primitives (MP)~\cite{Liu2023marching} and EMS~\cite{liu2022EMS}, and topology optimization approach Flexicubes (FC)~\cite{shen2023flexicubes}, MeshAnything (MA)~\cite{chen2024meshanything} and PrimitiveAnything (PA)~\cite{ye2025primitiveanything}.

\cameraReady{For PA, MA, MP, EMS, NVD, FC, CAPRI, D$^{2}$CSG, and DiffCSG, SDF values exported from RNb-NeuS are used as the input. For 2DGS+MC and 2DGS+CSG, we use the SDF exported from 2DGS as the input.}
\xiaoxu{
% We employ the normal maps estimated by StableNormal \cite{ye2024stablenormal} as supervision for both our method and RNb-NeuS.
Normal maps estimated by StableNormal~\cite{ye2024stablenormal} are employed as supervision for both our method and RNb-NeuS.
}
% For Flexicubes, PA, MA, MP, and EMS, we use the SDF values exported from RNb-NeuS as the input.

% \input{tables/metric.tex}

\vspace{-2mm}
\subsection{Reconstruction from Multi-view Images}
We report Chamfer Distance (CD) averaged over $180$ examples in Table~\ref{table:comparison_cd_face}. 
\MethodNameShort~achieves lower reconstruction errors compared to state-of-the-art approaches. 
RNb-NeuS~\cite{rnbneus}, \cameraReady{2DGS~\cite{huang20242dgs}} and nvdiffrec~\cite{Munkberg_2022_CVPR}, which extract 3D meshes from SDF grids using Marching Cubes~\cite{marching_cubes} or Marching Tetrahedrons~\cite{treece1999regularised}, tend to produce over-tessellated surfaces with limited geometric regularity. 
Flexicubes~\cite{shen2023flexicubes} recovers structured geometry, but grid-based mesh extraction can degrade fine details relative to meshes generated from primitives. 
Primitive-based methods, including Marching Primitives~\cite{Liu2023marching}, 
\cameraReady{
Capri-Net~\cite{yu2022capri}, D$^{2}$CSG~\cite{yu2023d},
DiffCSG~\cite{yuan2024diffcsg},
}
% DBW~\cite{monnier2023differentiable}, 
EMS~\cite{liu2022EMS}, MeshAnything~\cite{chen2024meshanything}, and PrimitiveAnything~\cite{ye2025primitiveanything}, achieve compact and interpretable meshes; however, the reconstruction quality depends on the underlying SDF and can be affected by noise and data loss.

We show qualitative results in Figure~\ref{fig:comparison_shapenet}.
While most methods successfully reconstruct the overall geometry, their outputs exhibit varying levels of fidelity and compactness. 
RNb-NeuS~\cite{rnbneus}, \cameraReady{2DGS~\cite{huang20242dgs}} and nvdiffrec~\cite{Munkberg_2022_CVPR}, which extract 3D meshes from SDF grids using Marching Cubes~\cite{marching_cubes} or Marching Tetrahedrons~\cite{treece1999regularised}, tend to produce overly tessellated surfaces that fail to respect underlying geometric regularities.
Flexicubes~\cite{shen2023flexicubes} demonstrates an ability to recover structured geometry; however, its reliance on grid-based mesh extraction can further degrade fine details compared to meshes directly generated via Marching Cubes. 
Methods such as MP~\cite{Liu2023marching}, MA~\cite{chen2024meshanything}, PA~\cite{ye2025primitiveanything},
\cameraReady{
CAPRI~\cite{yu2022capri},
D$^{2}$CSG~\cite{yu2023d},
DiffCSG~\cite{yuan2024diffcsg}
}
improve mesh compactness, but their reconstruction quality remains highly sensitive to the imperfections of the underlying SDF, which often manifest as surface artifacts and redundancies in the reconstructed meshes.
% Due to its limited shape-representation capacity, DBW~\cite{monnier2023differentiable} attempts to capture geometric details through textures, which prevents it from recovering a truly high-fidelity mesh.
% In particular, noise and geometric information loss in the SDF grid of RNb-NeuS introduce surface artifacts and redundancies in the resulting meshes.
% \zhongmin{As a learning-based method, }BSPNet's reliance on BSP trees of planes often results in blocky shapes with irregular triangulation patterns.
\zhongmin{In contrast, \MethodNameShort~reconstructs compact, high-fidelity meshes that closely preserve the shape and topological characteristics of artist-designed models.}
% In contrast, our method reconstructs compact, high-fidelity 3D meshes that closely resemble the shape and topological quality of artist-designed models.
\zhongmin{Additional qualitative and per-category quantitative results are provided in the supplementary material.}
% Please refer to the supplemental material for additional comparisons and per-category quantitative results.

% \xiaoxu{We also compare our approach with BSP-Net, a single-view reconstruction approach。  }

% \input{tables/comparison_mesh.tex}

% TODO: result analysis.
\vspace{-1mm}
\subsection{User Study on Geometry Editability}
\zhongmin{We conducted a user study to evaluate geometric editability compared to baselines. Twelve representative cases were randomly selected, and 36 participants with 3D mesh editing experience rated editability and structural controllability scored from 1 to 10.} 
\zhongmin{The results, summarized in second row of Table~\ref{table:comparison_cd_face}, show that \MethodNameShort~produces compact and structured geometric representations with significantly higher editability scores compared to prior methods. 
Further details are provided in the supplementary material.}
% To assess the geometric editability of our approach, we conducted a user study comparing it with seven representative baseline methods. Twelve representative cases were randomly selected for evaluation, and 33 participants rated the editability and structural controllability of each result on a \xiaoxu{score} from 1 to 10.} \xiaoxu{All participants are at least 18 years old with 3D mesh editing experience.
% The geometry editability scores are shown in Table~\ref{table:comparison_cd_face}.}
% \zhongmin{The user study demonstrates that our method reconstructs compact and structured geometric representations, showing a significant advantage in geometric editability over state-of-the-arts.} \xiaoxu{Please refer to the supplementary for the detail on the user study.}

\vspace{-4mm}
\begin{table}[htbp]
\scriptsize
\centering
\setlength\tabcolsep{4pt}
\begin{tabular}{l||ccccc}
\textbf{Method} & \textbf{\#V(k)}$\downarrow$ & \textbf{\#F(k)}$\downarrow$ & \textbf{NC Loss}$\downarrow$ & CD$\downarrow$ & EditScore$\uparrow$
\\
\toprule
% {Ours}   &   1.54 &   0.79 &    7.73&     7.94&     8.29
% \\NVD & FC & PA & MP & MA & EMS & DBW
{RNS~\cite{rnbneus}}    & 124.41 &  57.44 &     37.45&     10.71&    4.77
\\
{NVD~\cite{Munkberg_2022_CVPR}}    &  15.40 &  13.98 &     56.19&     10.63&     5.47
\\
{FC~\cite{shen2023flexicubes}}     &  16.67 &   8.33 &     16.11&     13.32&     4.98
\\
{PA~\cite{ye2025primitiveanything}}     &  21.51 &  12.61  &     4.48&     11.57&     4.59
\\
{MP~\cite{Liu2023marching}}     &   3.38 &   1.75 &     12.42&     12.58&     4.88
\\
{MA~\cite{chen2024meshanything}}     &   0.55 &   0.29 &     4.00&     10.12&     5.12
\\
{EMS~\cite{liu2022EMS}}    &   2.40 &   1.22 &     9.02&     16.64&     3.92
\\
% {DBW~\cite{monnier2023differentiable}}    &   0.24 &   0.17 &     3.53&     15.71&     3.09
% \\
{2DGS~\cite{huang20242dgs} + MC}   & 151.71 & 75.95 & 53.75 & 12.06 & 3.75
\\
{2DGS~\cite{huang20242dgs} + CSG}  &  44.83 & 18.10 & 16.66 & 12.63 & 4.41
\\
{DiffCSG~\cite{yuan2024diffcsg}}     &  10.24 &  5.20 &  8.20 & 11.27 & 5.50
\\
{CAPRI~\cite{yu2022capri}}       &  29.07 & 12.64 & 11.55 & 11.28 & 4.98
\\
{D$^{2}$CSG~\cite{yu2023d}}  &  24.27 & 11.62 & 13.21 & 12.12 & 4.73
\\
\toprule
{\textbf{Ours}}         &   1.54 &  0.79 &  7.73 &  7.94 & 8.29 
% Our new score: (${8.43^{*}}$)
\\
\bottomrule
\end{tabular}
% \vspace{-3mm}
\caption{Quantitative comparison and user evaluation of \MethodNameShort~(ours) with 13 representative methods.}
\label{table:comparison_cd_face}
\vspace{-4mm}
\end{table}

\subsection{Ablation Study}
% \paragraph{Regularizations on normal supervision.}
% We conduct an ablation study on the normal loss. Normal loss is helpful in our 3D reconstruction, especially when a large area shares the same RGB color in the RGB image and the lighting condition is not good enough to visualize the 3D shape. 
% As shown in Figure~\ref{fig:ablation_normal} (b), when $\lambda_{normal} = 0$, the main structure of the sofa can be reconstructed. However, the renderer cannot fully capture the detail of the sofa back, resulting in a non-flat surface.
% We use the advanced normal prediction approach, StableNormal~\cite{ye2024stablenormal}, to predict the normal maps. In regions with non-consistent multiview normals as shown in Figure~\ref{fig:ablation_normal_error} (b), our renderer is able to tolerate the error and reconstruct the 3D geometry, as shown in Figure~\ref{fig:ablation_normal_error} (c).
% \vspace{-4mm}
\paragraph{Ablation on regularization.}
% By setting $\mathcal{L}_{e}=0$, the $\transparency$ for effective primitives might distribute in $(0, 0.5)$, thus resulting in missing primitive with $T_{export} = 0.5$ as shown in Figure~\ref{fig:ablation_regularization} (b) and (c).
Comparing Figures~\ref{fig:ablation_regularization} (b) and (c), we observe that setting $\mathcal{L}_{sp}=0$ results in redundant primitives, rather than encouraging the model to use a sparse set.
Similarly, comparing Figures~\ref{fig:ablation_regularization} (e) and (f), setting $\mathcal{L}_{max}=0$ leads to multiple primitives with overlaps, instead of representing the geometry with a single primitive.
\vspace{-8mm}
\xiaoxu{
\paragraph{Ablation on PSQ-only training.}
Figure~\ref{fig:ablation_regularization} (g) and (h) compares the reconstruction using only PSQ versus using both PSQ and NSQ.
Incorporating NSQ allows the model to subtract local volumes and refine fine-scale structures while maintaining analytical regularity.
}
% \vspace{-4mm}
% \paragraph{Ablation on Training Parameters.}
% We conduct an ablation study on the number of initial primitives $K$, transparency threshold $\alpha$, and dual-primitives scale threshold $t_{\alpha}$.
% As shown in Figure~\ref{fig:ablation_k} (a) - (c), initializing with fewer primitives ($K$) limits the model’s deformation capability, resulting in lower-quality reconstructions. In addition, setting $\alpha$ or $t_{\alpha}$ too high causes excessive pruning, which negatively impacts the final mesh quality.

% Comparing Figure~\ref{fig:ablation_regularization} (b) and (c), by setting $\mathcal{L}_{sharpness}=0$, the renderer will keep the redundant surfaces, instead of learning a validity space as sparse as possible.

\vspace{-1mm}
\section{Conclusion and Future Work}
We introduce \MethodNameShort, a novel algorithm for reconstructing artist-style 3D meshes from multi-view images.
\zhongmin{Our approach employs a primitive-based representation that is both compact and differentiable, enabling seamless integration with volume rendering pipelines. }
% Our approach employs a primitive-based representation that is compact and differentiable, thus supporting integration with the volume rendering pipeline.
\zhongmin{Extensive experiments demonstrate that \MethodNameShort~outperforms current state-of-the-art methods in reconstruction accuracy, compactness, and topological quality across diverse object categories. }
% This method outperforms current state-of-the-art techniques in accuracy, compactness, and topology quality as demonstrated across different types of objects. 
\zhongmin{We believe this advancement will have a significant impact on the generation of 3D assets for gaming and the metaverse, and has the potential to inspire further research in areas such as physics simulation and collision detection.}
% We believe this advancement will significantly impact the generation of 3D assets for gaming and the metaverse, while also inspiring further research in physics simulation and collision detection.
% Limitation: resolution
Nevertheless, our algorithm has certain limitations. It struggles to accurately capture object parts that are thinner than the voxel grid resolution, though this can be mitigated by increasing the grid resolution or thickening the affected regions. 
\cameraReady{Another limitation is primitive overlap. Although the regularization terms substantially reduce overlap, minor edge-level or layout-induced intersections may still occur. These small overlaps do not noticeably affect reconstruction fidelity or downstream editability.}
Furthermore, the method is primarily tailored for industrially designed objects with well-defined structures and demonstrates reduced performance on real-world objects with irregular or less structured geometries.
\vspace{-3mm}
\cameraReady{
\paragraph{Acknowledgements.}
This work was supported by the National Natural Science Foundation of China (No. 62322210, 62561160115). We would like to thank the anonymous reviewers for their constructive feedback and insightful suggestions, which helped improve the quality of this work. We also thank Haocheng Yuan and Yixuan Li for their generous assistance in running the comparison experiments for DiffCSG and PrimitiveAnything.
}

{
    \small
    \bibliographystyle{ieeenat_fullname}
    \bibliography{main}
}

\end{document}